\documentclass[runningheads]{llncs}

\usepackage{eccv}

\usepackage{eccvabbrv}

\usepackage{graphicx}
\usepackage{booktabs}

\usepackage{multirow}
\usepackage{algorithm}
\usepackage{alphalph}
\usepackage{algorithmic}
\usepackage{bbding}
\usepackage{booktabs}
\usepackage{makecell}

\usepackage[accsupp]{axessibility}

\usepackage{hyperref}

\usepackage{orcidlink}
\newcommand{\gray}[1]{{\color[rgb]{0.753,0.753,0.753} #1}}

\begin{document}

\title{PSALM: Pixelwise SegmentAtion with Large Multi-Modal Model}

\author{Zheng Zhang\inst{1,2}\thanks{Equal contribution} \and Yeyao Ma\inst{1}$^\star$ \and Enming Zhang\inst{1}$^\star$ \and Xiang Bai\inst{1}}

\authorrunning{Z. Zhang, Y. Ma, E. Zhang, X. Bai}

\institute{Huazhong University of Science and Technology \and Microsoft Research Asia}

\maketitle

\begin{abstract}
PSALM is a powerful extension of the Large Multi-modal Model (LMM) to address the segmentation task challenges. To overcome the limitation of the LMM being limited to textual output, PSALM incorporates a mask decoder and a well-designed input schema to handle a variety of segmentation tasks. This schema includes images, task instructions, conditional prompts, and mask tokens, which enable the model to generate and classify segmentation masks effectively. The flexible design of PSALM supports joint training across multiple datasets and tasks, leading to improved performance and task generalization. PSALM achieves superior results on several benchmarks, such as RefCOCO/RefCOCO+/RefCOCOg, COCO Panoptic Segmentation, and COCO-Interactive, and further exhibits zero-shot capabilities on unseen tasks, such as open-vocabulary segmentation, generalized referring expression segmentation and video object segmentation, making a significant step towards a GPT moment in computer vision. Through extensive experiments, PSALM demonstrates its potential to transform the domain of image segmentation, leveraging the robust visual understanding capabilities of LMMs as seen in natural language processing. Code and models are available at \url{https://github.com/zamling/PSALM}. 
\keywords{Segmentation \and Large Multimodal Model \and Visual Language}
\end{abstract}

\label{sec:intro}
\begin{figure}[!t]
  \centering
  \includegraphics[width=0.9\textwidth]{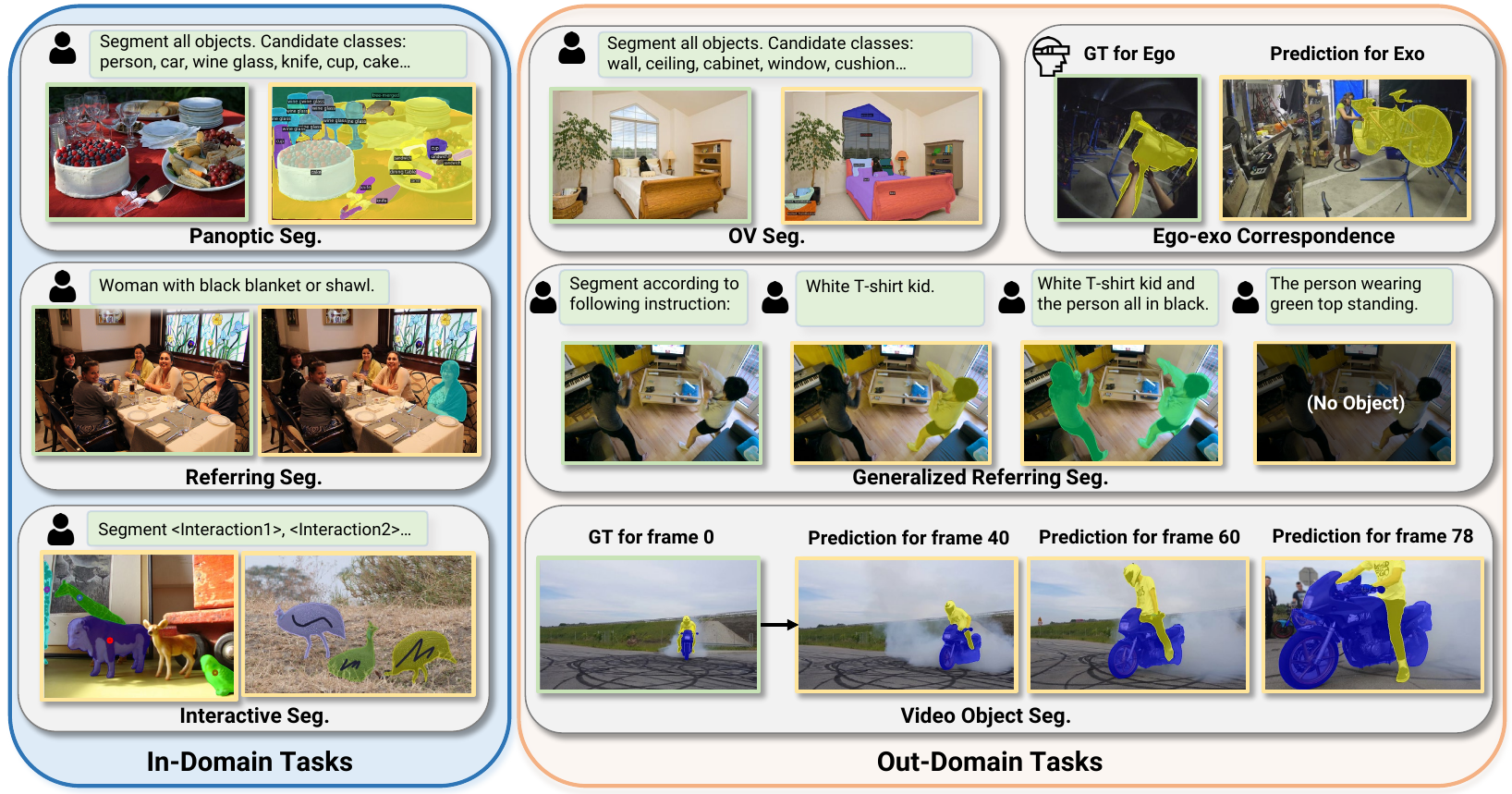}
  \caption{PSALM has capability to handle multiple segmentation tasks in only one single model. We visualize some tasks, including Panoptic segmentation in COCO~\cite{coco}; Open-Vocabulary instance segmentation in ADE20K~\cite{ade20k}; Interactive segmentation in COCO-Interactive; Referring segmentation in RefCOCO~\cite{refcoco}; Generalized referring segmentation in gRefCOCO~\cite{grefcoco}; Ego-exo correspondence in Ego-Exo4d~\cite{egoexo}; Video object segmentation in DAVIS2017~\cite{davis}.}
  \label{fig:teaser}
\end{figure}
\section{Introduction}
Large multi-modal model (LMM) has ignited the dawn of the vision GPT~\cite{GPT3} moment by making ground-breaking progress in various advanced visual understanding tasks by compressing image and language information into a single auto-regressive model. 
However, a major obstacle on the road to achieving vision GPT is that current LMM can only perform text outputs, making it challenging to address the pixel-level image understanding problem directly, \textit{i.e.}, image segmentation, which is one of the most critical tasks in computer vision.

Behind the obstacles are many challenges. First, the default output of the LMM is discrete tokens, and there is no apparent way to generate masks directly. Second, the variety of image segmentation tasks requires different forms of inputs and outputs. For example, semantic segmentation needs support inputs with different categories. Instance segmentation demands generate object IDs and the class confidence score for each object. Referring segmentation is supposed to have a language sentence as input and interactive segmentation has more varied inputs than the other tasks, which can be points, scribbles, bounding boxes, or masks. Third, unifying different segmentation tasks with a shared weight model is also challenging because different tasks require varied capabilities.

In this work, we propose a method named PSALM (\textbf{P}ixelwise \textbf{S}egment\textbf{A}tion with \textbf{L}arge Multi-Modal \textbf{M}odel) that aims to address the above challenges and extend the capabilities of LMM from text-output tasks to general segmentation tasks (Fig.~\ref{fig:teaser} shows representative tasks). Specifically, PSALM externalizes a mask decoder on the top of LMM and designs a flexible input schema to unify different segmentation tasks into a single model.

The input schema consists of four different parts: images, task instruction prompt, condition prompt, and a set of mask tokens, where the instruction prompt is a text sentence describing the task itself, condition prompt contains the additional necessary information to solve the task, either in terms of category names, sentence or visual features, and mask tokens are a set of learnable embeddings. All these inputs are fed into the LMM, and the resulting output mask tokens are further used as input by the mask generator to present mask proposals. 
Apart from producing the mask proposals, it is also necessary to predict the class of each segmentation mask or estimate a confidence score, which can be achieved by using the output embedding of the condition prompt as the classifier weights to classify each mask proposal. In practice, we categorize conditions into category condition, sentence condition, and visual-prior condition, and present the corresponding methods to build the classifier weights according to the properties of each type of condition. 

Some other methods, represented by LISA~\cite{lisa}, also aim to use LMM for segmentation tasks. However, these methods are usually designed for referring segmentation and fail to justify their ability to solve generalized segmentation tasks (see Tab.~\ref{tab:capability_comparison}).
In contrast, thanks to the generality and flexibility of the proposed architecture, PSALM can not only solve a variety of segmentation tasks but also be able to joint train on different tasks, which makes the model task generalizable while allowing the model to take full advantage of the intrinsic connections of different datasets/tasks to achieve better performance. 
Specifically, with the joint training of COCO Panoptic Segmentation~\cite{coco}, RefCOCO~\cite{refcoco}/RefCOCO+/RefCOCOg~\cite{refcocog}, and COCO Interactive, we observe a significant performance improvement compared to training at different tasks individually, and therefore result in even better performance than other task-specific methods. On referring segmentation tasks, we outperform other LLM-based pixel reasoning methods (\textit{e.g.}, LISA, PixelLM~\cite{pixellm} and GSVA~\cite{xia2023gsva}) on RefCOCO, RefCOCO+, and RefCOCOg, and it worth noting that we only use Phi-1.5 1.3B model~\cite{phi15} while others adopt Vicuna-7B~\cite{vicuna} or LLama2-13B model~\cite{llama2}.

The flexible design of architecture and input schema, multi-task joint-training, and the strong visual understanding capability of LMM not only make PSALM perform well on trained in-domain tasks but also enable generalizability to out-of-domain tasks in a zero-shot manner, \textit{i.e.,} directly dealing with unseen tasks without additional training. We test on three tasks: open-vocabulary segmentation, generalized referring expression segmentation, and video object segmentation. PSALM achieves promising zero-shot performance on these tasks. We believe this task-level generalizability is crucial, which is one of the key properties of the large language model for its success in natural language processing.

Through extensive experiments on a variety of segmentation tasks, we show that presented PSALM has strong potential to address general image segmentation tasks and exhibits a certain degree of task generalization capability as LLM does in NLP. We believe this work facilitates the realization of the GPT moment in computer vision.
\begin{table}[t]
\centering
\footnotesize
\caption{Capability for different methods. Our proposed PSALM can handle more segmentation tasks than other LLM-centric methods. LLM-centric methods can also deal with text generation tasks, which is hard for most vision-centric methods.}
\scalebox{0.8}{
\begin{tabular}{l|lclclclc}
\toprule
                           & Methods    & Generic Seg.              &  & Referring Seg.            &  & Interactive Seg.          &  & OV Seg.                   \\ \midrule
\multirow{5}{*}{{Vision centric}}   & Mask2Former~\cite{mask2former} & \Checkmark &  &                           &  &                           &  &                           \\
                           & ODISE~\cite{odise}      & \Checkmark &  &                           &  &                           &  & \Checkmark \\
                           & UNINEXT~\cite{uninext}    & \Checkmark &  & \Checkmark &  & \Checkmark &  &                           \\
                           & SEEM~\cite{seem}       & \Checkmark &  & \Checkmark &  & \Checkmark &  & \Checkmark \\
                           & OMG-Seg~\cite{omgseg}    & \Checkmark &  &                           &  & \Checkmark &  & \Checkmark \\ \midrule
\multirow{4}{*}{{LLM centric}} & LISA~\cite{lisa}       &                           &  & \Checkmark &  &                           &  &                           \\
                           & GLAMM~\cite{glamm}      &                           &  & \Checkmark &  &                           &  &                           \\
                           & PixelLM~\cite{pixellm}    &                           &  & \Checkmark &  &                           &  &                           \\ 
                           & PSALM (Ours)      & \Checkmark &  & \Checkmark &  & \Checkmark &  & \Checkmark \\ \bottomrule
\end{tabular}
}
\label{tab:capability_comparison}
\end{table}
\section{Related works}
\subsection{Large Multimodal Models}
With the release of GPT-V~\cite{GPT-V} and Gemini~\cite{Gemini}, more attention and efforts from open-source and research communities are shifting from large language models (LLM) to large multi-modal models (LMM). 
LLaVA~\cite{llava}, BLIP-2~\cite{blip2}, and Flamingo~\cite{flamingo} are three representative works, where the core idea of both LLaVA and BLIP-2 is to map visual features into the input space of LLM to implement multi-modal capabilities, while Flamingo employs deeper feature fusion in the intermediate layers of LLM. 
Some works, such as Kosmos-2~\cite{kosmos_2}, Shikra~\cite{chen2023shikra}, and Ferret~\cite{you2023ferret}, further introduce object localization tasks into the LMM, while others, such as Emu~\cite{emu}, CogVLM~\cite{wang2023cogvlm}, and DreamLLM~\cite{dong2023dreamllm}, focus on how to integrate visual generation into the LMM. 
However, these above methods are mainly designed for text output tasks or image generation and cannot directly deal with pixel-level understanding tasks such as image segmentation, which is different from ours and we can base on these models.
\subsection{Pixel Reasoning with LMM}
Similar to our goal, some existing works attempt to enable LMMs to generate segmentation masks. LISA~\cite{lisa} is a pioneering work that uses a special seg token to aggregate information of a given sentence and use it as a prompt embedding in a SAM decoder to predict the segmentation mask. u-LLaVA~\cite{ullava} further supports object grounding tasks on the basis of LISA, and NExT-Chat~\cite{nextchat} introduces richer inputs, such as bounding boxes. Furthermore, since LISA can only deal with a single object, many subsequent works that attempt to address the multi-object case, such as GLaMM~\cite{glamm}, PerceptionGPT~\cite{pi2023perceptiongpt}, PixelLM~\cite{pixellm}, GSVA~\cite{xia2023gsva}, and LISA++~\cite{lisa++}, all of which share the basic idea of introducing a seg token for each sentence describing a different object, and except PixelLM, all other methods are based on SAM decoder~\cite{sam}.

Although all these methods can generate masks, they are primarily designed for reference segmentation. In contrast, our method is designed for generalized segmentation tasks, which have diverse input and output requirements. In addition, the difference in goals also brings technical discrepancies: these methods use language models to directly generate the final segmentation masks, while our approach is closer to Mask2Former~\cite{mask2former}, which first generates mask proposals and then classifies the masks.
\begin{figure}[!t]
  \centering
  \includegraphics[width=.9\textwidth]{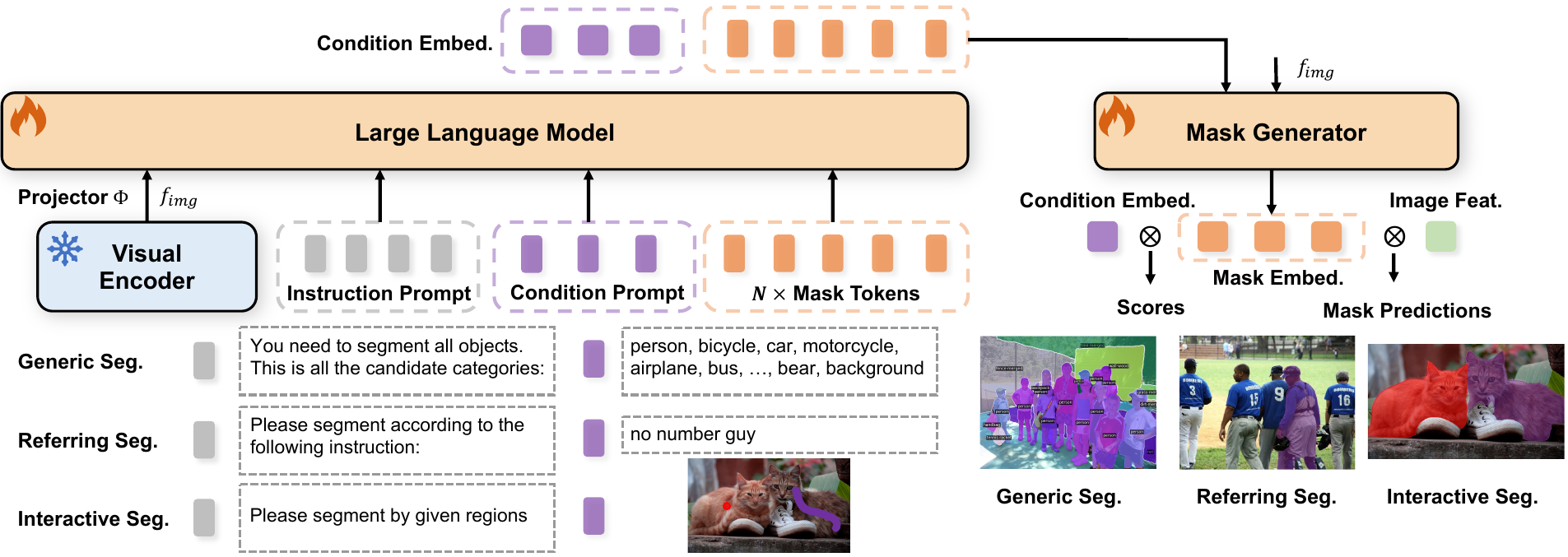}
  \caption{PSALM architecture overview.}
  \label{fig:overview}
\end{figure}
\subsection{Unified Segmentation Framework}
Another class of relevant explorations~\cite{uninext,seem,omgseg,glee} studies how to unify different segmentation tasks into a single framework. For instance, Mask2Former~\cite{mask2former} presents a unified architecture of generic segmentation~\footnote{Generic segmentation includes semantic segmentation, instance segmentation, and panoptic segmentation.} but with different models.  OneFormer~\cite{jain2023oneformer} further integrates these three tasks within a single model. UNINEXT~\cite{uninext} aims to unify instance-perception tasks and introduce text inputs and thus it can handle referring segmentation. X-Decoder~\cite{xdecode} presents a flexible decoder architecture that can support generic segmentation, referring segmentation, retrieval, and image captioning. SEEM~\cite{seem} designs a generic encoder-decoder to unify different segmentation tasks, where the encoder is used for projecting image, text, and human inputs into a joint visual-semantic space, and the decoder is used for mask prediction. 
However, all these works are not based on LMM, instead, they are mostly vision-centric models, \textit{i.e.}, usually designed for visual tasks only and thus cannot address language tasks very well.

\section{Methods}

Fig.~\ref{fig:overview} provides an overview schematic of PSALM, which consists of a large multimodal model (LMM), a mask generator, and a flexible input schema designed for general segmentation tasks. The input schema has four different types of inputs: image, task instruction prompt, condition prompt, and a set of mask tokens. LMM processes the input tokens and the output embedding of mask tokens is further fed into the mask generator to generate masks. In the following, we will introduce our approach in detail.
\subsection{Large Multimodal Model and Input Schema}
PSALM is built on large multimodal models (LMM), and there are many different LMM architectures, such as LLaVA~\cite{llava}, BLIP~\cite{blip}, and Flamingo~\cite{flamingo}. Here, we adopt the design of LLaVA because of its proven performance and simplicity, but the other LMM architectures are also compatible with our approach without any theoretical difficulties.

The LMM used in our work has a visual encoder and pre-trained large language model (LLM). The two models are connected by a lightweight vision-language alignment model, which is a $3\times3$ convolution layer followed by a linear layer. 
The official LLaVA model uses a frozen CLIP model~\cite{CLIP} as a visual coder, whose features lack the fine-grained information that is required for segmentation tasks~\cite{SimBaseline}. Therefore, we train a customized LLaVA model by using the Swin Transformer~\cite{Swin}, and limited by resources, we additionally replace the LLM from the Vicuna 7B model~\cite{vicuna} to a smaller Phi-1.5 1.3B model~\cite{phi15}. Here, we applied only the first visual-language alignment stage of LLaVA by following its default settings. In our ablations, we found the alignment stage is crucial for open-vocabulary segmentation and referring segmentation tasks (Tab.~\ref{tab:vl_align}).

Different segmentation tasks need different forms of inputs and outputs, which motivates us to present a flexible input schema to unify various requirements. In addition to the input image used in the visual encoder, our input schema has three other different types of inputs: task instruction prompt, condition prompt, and a set of mask tokens. We will introduce each of them and summarize the prompts used for all different tasks in the Appendix.

\noindent\textbf{Task Instruction Prompt.} The task instruction prompt is usually a text sentence describing and specifying the model's task. For example, in panoptic segmentation and open-vocabulary segmentation, the task instruction can be \textit{“You need to segment all objects. This is all the candidate categories.”} and in referring segmentation, the instruction can be \textit{“Please segment according to the following instruction.”}

\noindent\textbf{Condition Prompt.} Some tasks require additional information to perform, \textit{e.g.}, panoptic segmentation needs specifying the candidate set of categories to be segmented, and interactive segmentation needs interactive inputs. The condition prompt is designed for these tasks. In addition to providing information, condition prompt also plays an important role in predicting categories or estimating confidence scores for each segmentation mask. In Sec.~\ref{sec:condition_prompt}, we will discuss the design of condition prompts for different tasks in detail.

\noindent\textbf{Mask Token.} The LLM is designed for text output and cannot directly generate segmentation masks. To bypass this challenge, we append a set of mask tokens after other inputs, and then these mask tokens are decoded to segmentation masks by a mask generator (will be introduced in Sec.~\ref{sec:mask_generator}). This design is inspired by Mask2Former~\cite{mask2former}, with the difference that the mask tokens in Mask2Former are used directly in the mask generator, whereas the mask tokens in our approach are first updated by the LMM and then used in the mask generator, and we found our approach leads better performance in practice (see Tab.~\ref{tab:mask_query}).

Some works, such as LISA~\cite{lisa} and PixelLM~\cite{pixellm}, take similar seg tokens as input and employ a decoder to generate the masks. However, our objective is fundamentally different: in LISA and PixelLLM, seg tokens are used to generate the final prediction, while we generate mask proposals and further classify them based on condition prompts. Compared to the design of LISA and PixelLLM, our approach is more flexible and adaptable to more tasks, yet decoupling mask generation and classification is more efficient (see Tab.~\ref{tab:decouple} and Tab.~\ref{tab:mask_proposals}).

\subsection{Mask Generator}
\label{sec:mask_generator}
The mask generator predicts the mask and their category probabilities from three inputs: a multi-level visual features $\{v_{l}\}_{l=1}^{L}$, a set of mask tokens $\{q_i\}_{i=1}^{N}$, and a set of condition embeddings $\{c_{k}\}_{k=1}^{K}$. It can be formally defined as:
\begin{equation}
    \small
    \{(m_{i}, p_{i})\}_{i=1}^{N} = \texttt{MaskGenerator}(\{v_{l}\}_{l=1}^{L}, \{q_{i}\}_{i=1}^{N}, \{c_{k}\}_{k=1}^{K}),
\end{equation}
where $m_{i}\in \mathbb{R}^{H\times W}$ is the i-th predicted segmentation mask and $p_{i}\in \mathbb{R}^{K}$ is the corresponding category probability. 

In practice, the multi-level visual features are the internal output features of the Swin visual encoder used in LMM, and the design of the mask generator follows Mask2Former, which employs multi-scale deformable attention as a pixel decoder and a transformer-based mask decoder to generate segmentation masks. The class of each mask is predicted by the condition embedding $\{c_{k}\}$, which is basically obtained from the output of the condition prompt, the obtaining method is slightly different for different types of conditions.

\begin{figure}[!t]
  \centering
  \includegraphics[width=0.9\textwidth]{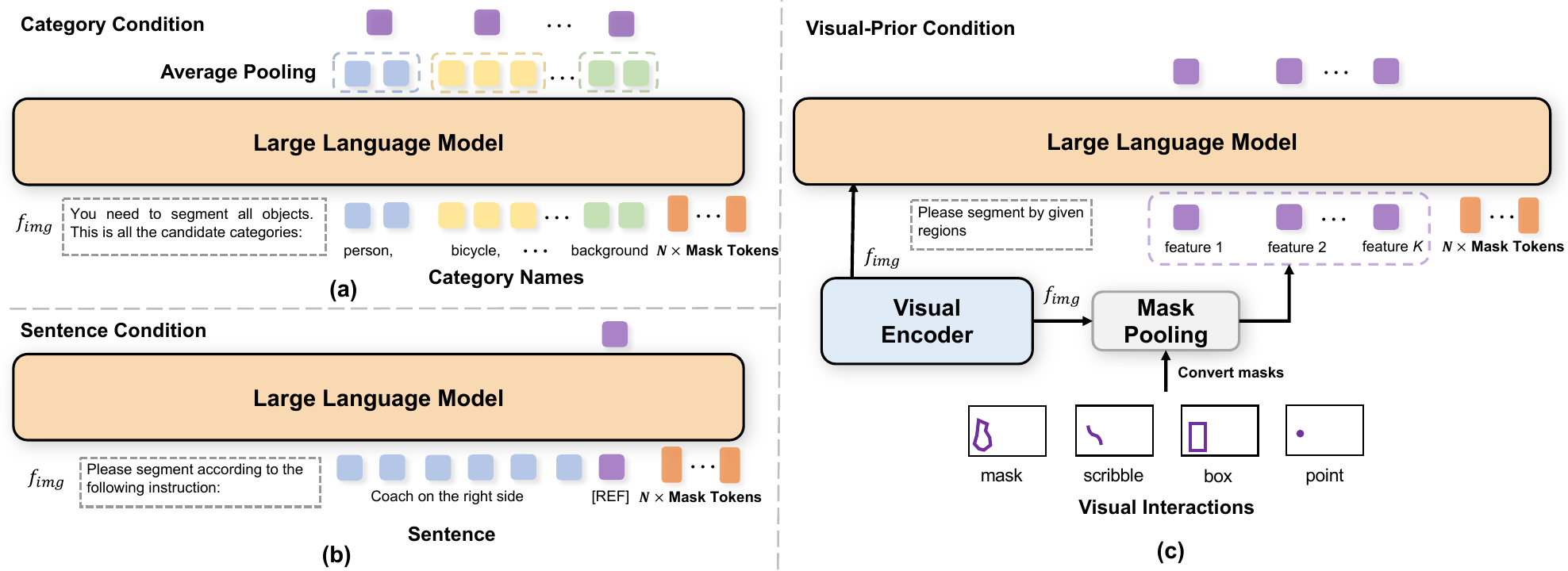}
  \caption{Detailed processing for different condition prompts. (a) shows the processing for category condition. (b) shows the processing for sentence condition. (c) shows the processing for visual-prior condition.}
  \label{fig:taskdetail}
\end{figure}
\subsection{Design of Condition Prompts}
\label{sec:condition_prompt}
In our approach, the conditional prompt serves two important purposes: First, it provides the necessary information required to solve the task; Second, we use the output embedding of the conditional prompt in the LLM as classifier weights to predict the class of each segmentation mask. The design of the conditional prompt is closely tied to the type of task, and based on the information required for different tasks, we summarize three condition types: category condition, sentence condition, and visual-prior condition.

\noindent\textbf{Category Condition} This condition type is used for tasks that need specifying a set of categories to be segmented, such as semantic segmentation, instance segmentation, and panoptic segmentation, and often needs to predict the class probability of each segmentation mask. Specifically, given a set of category names, we join them into a sentence by a comma separator, \textit{e.g.,}, given three categories: person, bicycle, and car, the joint sentence is \textit{"person, bicycle, car"}. 

The joint sentences are then proceeded by LMM to get the output embeddings, which can be further used to classify the predicted segmentation masks. Specifically, for each category, we select its corresponding output embeddings and apply the \texttt{avg\_pooling} over them to obtain a condition embedding $c \in \mathbb{R}^{D}$, where $D$ is the embedding dimension, and thus condition embedding for all categories is a set $\{c_{k}\}_{k=1}^{K}$, where $K$ is the number of categories (see Fig.~\ref{fig:taskdetail} (a)). This embedding set can be used by the mask generator to predict the class.

\noindent\textbf{Sentence Condition} This condition is usually used for referring segmentation. Unlike the category condition, where category names are usually short, sentences are much longer, and not every word in the sentence is useful, so the \texttt{avg\_pooling} is not the optimal choice here. Instead, we introduce a special \texttt{[REF]} token, which is appended after the condition sentence as an anchor to aggregate useful information, and the output embedding of \texttt{[REF]} token, \textit{i.e.,} the output features of LMM on the location of \texttt{[REF]} token, is used as condition embedding $c$ and used by mask generator, as shown in Fig.~\ref{fig:taskdetail} (b).

\noindent\textbf{Visual-Prior Condition} We formulate most interaction (\textit{e.g.,} point, mask, box, or scribble) used in interactive segmentation tasks as the visual-prior condition. Taking the scribble as an example, we first generate a binary scribble curve with a width of 5 pixels on a binary map, whose size is the same as the input image of the Swin vision encoder, and then apply the binary map to the output features of the vision-language alignment model, performing average mask pooling by upsampling the output feature map to the size of the binary map. The average pooled feature is used as the visual-prior condition and fed into LMM. 
If there are multiple interactions, the process is repeated for each, yielding multiple pooled features as inputs, each feature separated using a strategy akin to the category condition. 
For other types of interaction, we adopt similar approaches. Specifically, for box or mask, we directly apply the pooling operator by treating them as binary masks, and for point, we bold it to a 10-pixel circle and then apply the mask pooling. 
After that, we use the output embedding of the visual-prior condition as the classifier weight in the mask generator to estimate the confidence of each mask proposal, as shown in Fig.~\ref{fig:taskdetail} (c).

\subsection{Training Objectives}
The training process of PSALM can be divided into two stages: In the first stage, we train the visual language alignment model following LLaVA setting, and freeze the vision encoder and LMM; In the second stage, we only freeze the vision encoder and fine-tune all other modules, including the mask generator. Similarly to Mask2Former, we use matching loss in the second stage training, \textit{i.e.}, we use bipartite matching to find the optimal assignments between the mask proposals and the ground truth masks by minimizing the matching loss and use these assignments to perform training. The loss has two terms: $\mathcal{L}=\mathcal{L}_{mask} + \mathcal{L}_{cls}$, where $\mathcal{L}_{mask}$ indicates the mask loss which comprises a pixel-level Binary Cross-Entropy (BCE) loss and Dice loss~\cite{DiceLoss}, while the $\mathcal{L}_{cls}$ indicates the category classification loss, and we use Cross-Entropy (CE) loss for category condition and BCE loss for other cases. More training details are in the Appendix.

\section{Experiments}
\subsection{Datasets and Tasks}
In our experiments, many different datasets and tasks are involved. In training, we used a two-stage training strategy. In the first stage, we use CC3M~\cite{cc3m} to align the vision features to the text input space. In the second stage, we train the model on four different tasks and corresponding datasets: generic segmentation(COCO Panoptic Segmentation~\cite{coco}), referring segmentation(RefCOCO/+/g~\cite{refcoco,refcocog}), interactive segmentation(COCO-Interactive) and the vision-language instruction task (LLaVA1.5 training data~\cite{llavav1_5}). Note that COCO-Interactive is our in-house dataset, as there is no well-established public dataset that supports all four interaction types (point, scribble, box, and mask); we will release this dataset, and its construction details are given in the Appendix.

The evaluation tasks are classified into in-domain tasks and out-of-domain tasks, according to if the evaluation task appears in the training. Specifically, we use three out-of-domain tasks in this work: generalized referring expression segmentation(gRefCOCO~\cite{grefcoco}), open-vocabulary segmentation (ADE20K~\cite{ade20k}, Cityscapes~\cite{cordts2016cityscapes}, Pascal Context~\cite{pc}, and Pascal VOC~\cite{pascal-voc-2010}) and video object segmentation (DAVIS-2017~\cite{davis}), more dataset details are given in the Appendix.

\subsection{Implementation Details}
Swin-B~\cite{Swin} is used as a visual encoder with a Phi-1.5 1.3B~\cite{phi15}, the architecture of the mask generator is the same as Maks2Former, the number of mask tokens is set to 100, and both Swin-B and mask generator are initialized from the pre-trained Maks2Former model weight. By default, we train PSALM for 56k iterations by distributing equal training iterations across each task in system-level comparison and 9k iterations in ablations. All training images are resized to $1024^2$ by padding the shorter side to keep the aspect ratio; the batch size is 64. AdamW optimizer is used and sets the initial learning rate to $4\times10^{-5}$ with cosine schedule and without weight decay. If not specified, the model is trained with a joint training setting and without additional task-specific fine-tuning. All experiments are run on 16$\times$V100 GPUs.

\subsection{Ablations}
We first ablate key designs and present behind insights in this section. To better show how different designs affect the performance on a wide range of tasks, we mainly report the results on three in-domain benchmarks: COCO Panoptic Segmentation (COCO-Pan), RefCOCO-val (RefCOCO), and COCO Interactive Segmentation with point inputs (COCO-Point), and one out-of-domain benchmark: open-vocabulary instance segmentation on ADE20K-150 (A150-OV).
\begin{table}[h!]
  \begin{minipage}{.6\textwidth}
  \centering
\footnotesize
\caption{Ablation on the design of mask tokens. w.LLM: use mask tokens as inputs of LLM. Prefix: place mask tokens at the front. Suffix: place mask tokens at the end.}
\renewcommand{\arraystretch}{0.9}
\scalebox{0.8}{
\begin{tabular}{cccccc}
\toprule
\multicolumn{2}{c}{Mask Tokens}               & COCO-Pan              & RefCOCO               & COCO-Point            & A150-OV              \\
\cmidrule{1-2}
w.LLM & Pos.   & PQ                    & cIoU                  & mIoU                  & mAP                   \\ \midrule
\Checkmark            & Suffix & 55.1                  & 76.1                  & 53.3                  & 9.3                  \\
\XSolidBrush          & -      & 54.8($\downarrow$0.3) & 74.3($\downarrow$1.8) & 53.1($\downarrow$0.2) & 8.2($\downarrow$1.1) \\
\Checkmark            & Prefix & 55.0($\downarrow$0.1) & 75.1($\downarrow$1.0) & 53.0($\downarrow$0.3) & 7.8($\downarrow$1.5) \\ \bottomrule
\end{tabular}
}
\label{tab:mask_query}

  \end{minipage}\hfill
  \begin{minipage}{.35\textwidth}
  \centering
\captionsetup{font=small}
\footnotesize
\caption{Effect of decouple design in COCO Semantic Segmentation. }
\scalebox{1.0}{
\centering
\begin{tabular}{ccc}
\toprule
Decouple & mIoU & fwIoU \\ \midrule
    \Checkmark         & 66.5          & 72.5          \\
    \XSolidBrush         & 42.7          & 35.0          \\ \bottomrule
\end{tabular}
\label{tab:decouple}
}

  \end{minipage}
\end{table}

\noindent\textbf{Design of Mask Tokens.}
In our approach, we use a set of mask tokens to predict the mask proposal. In practice, we have found that using mask tokens as inputs to the LLM leads to better performance than applying them directly to the mask generator, which is the default method of Mask2Former. Tab.~\ref{tab:mask_query} shows the results, where the direct use of mask tokens leads to a noticeable performance degradation in RefCOCO and A150-OV. 
We believe this is because using mask tokens as input leads to a better awareness of the information needed for the task and thus improves performance, which is essential for these two tasks. For better validation, we placed mask tokens before conditional prompts and task instruction prompts and found a similar performance drop to that of not using mask tokens in LMM, which further supports our hypothesis.

Compared to LISA and other methods that use seg token to generate final segmentation results directly, our mask proposal approach has three advantages: First, our design is more flexible and thus can be applied to a wider range of segmentation tasks, especially tasks that require predicting category or confidence scores; Second, our design decouples the mask prediction and classification, which alleviating the learning difficulties for some tasks. In Tab.~\ref{tab:decouple}, we study how the decouple design affects the semantic segmentation performance on COCO Semantic Segmentation\footnote{This benchmark is introduced by Mask2Former, which is composed by merging instances belonging to same class together in COCO Panoptic Segmentation.}, and we found our decouple design is significantly better\footnote{More experimental details are in Appendix.}. Third, the mask proposals allow multiple masks to be generated for a single instance, which makes the mask accuracy superior to solutions like LISA that only predict a single mask. Tab.~\ref{tab:mask_proposals} shows that using more mask proposals on RefCOCO gives a clear improvement over using a single mask.

\begin{table}[h!]
  \begin{minipage}{.4\textwidth}

\centering

\captionsetup{font=small}
\footnotesize
\centering
\caption{Effects of number of mask tokens in RefCOCO(cIoU). The results of LISA are listed as a reference.}
\scalebox{0.8}{

\begin{tabular}{cccccc}
\toprule
  \#Mask Token                  &  &  & val & testA & testB \\ \midrule
\gray{LISA} &          &  &   \gray{74.1}       &        \gray{76.5}   &   \gray{71.1}    \\
\midrule
1  &          &  &   75.3       &        78.0   &   72.2       \\
100   &           &  &   76.5      &   78.5       &   73.4       \\
\bottomrule
\end{tabular}
\label{tab:mask_proposals}

}
  \end{minipage}\hfill
  \begin{minipage}{.55\textwidth}
\centering
\footnotesize
\caption{Ablation on different designs for condition prompts.
}
\scalebox{0.8}{
\begin{tabular}{ccccc}
\toprule
\multicolumn{2}{c}{Condition}   & COCO-Pan  & RefCOCO  & A150-OV \\ \cmidrule{1-2}
Category & Sentence & PQ        & cIoU      & mAP        \\ \midrule
\texttt{avg\_pooling}          & \texttt{avg\_pooling}          & 55.1     & 75.3    & 9.2        \\
\texttt{[REF]}          & \texttt{[REF]}          & 54.9     & 76.1     & 8.4        \\
\texttt{avg\_pooling}          & \texttt{[REF]}          & 55.1      & 76.1    & 9.3        \\ \bottomrule
\end{tabular}
}
\label{tab:arch}

  \end{minipage}
\end{table}

\noindent\textbf{Design of Condition Prompts.}
Another key design in PSALM is the condition prompt, particularly the way we obtain the condition embeddings, which are used as classifier weights in the mask generator to predict the class of mask proposals. As described in Sec.~\ref{sec:condition_prompt}, for category condition, we use the \texttt{avg\_pooling} over output embeddings of each class name as the condition embedding, and for sentence condition, we adopt a \texttt{[REF]} token to aggregate useful information.

Tab.~\ref{tab:arch} shows the ablation, where we first tried to use the same design for all conditions and found that \texttt{avg\_pooling} performed slightly better on COCO-Pan, with a larger improvement on A150-OV, while \texttt{[REF]} worked better on RefCOCO. We further used different designs and found that the advantages of each design are preserved, and the best overall performance is achieved.

\begin{table}[h!]
  \begin{minipage}{.48\textwidth}
\centering
\footnotesize
\caption{Ablation on effect of vision-language alingment.}
\scalebox{0.65}{
\begin{tabular}{ccccc}
\toprule
\multirow{2}{*}{VL Alignment} & COCO-Pan              & RefCOCO           & COCO-Point             & A150-OV               \\
                              & PQ                    & cIoU                  & mIoU                   & mAP                   \\ \midrule
\Checkmark     & 55.1                  & 76.1                  & 53.3                   & 9.3                  \\
\XSolidBrush   & 54.9($\downarrow$0.2)  & 71.7($\downarrow$4.4)  & 53.0 ($\downarrow$0.3)  & 8.2($\downarrow$1.1) \\ \bottomrule
\end{tabular}
}
\label{tab:vl_align}
  \end{minipage}\hfill
  \begin{minipage}{.48\textwidth}
\centering
\footnotesize
\caption{Ablation on joint training.}
\scalebox{0.62}{
\begin{tabular}{c|cccc}
\toprule
\multirow{2}{*}{Ablation} & COCO-Pan  & RefCOCO & COCO-Point & A150-OV \\
                          & PQ           & cIoU        & mIoU   & mAP      \\ \midrule
Task Specific Train       & 55.6         & 76.5        & 62.2  & 7.0     \\
Joint Train               & 55.9         & 83.6        & 64.1  & 9.0     \\
$\Delta$                  & +0.3       & + 7.1       & + 1.9   & +2.0    \\ \bottomrule
\end{tabular}
}
\label{tab:joint}

  \end{minipage}
\end{table}

\noindent\textbf{Importance of VL-alignments.}
The visual-language alignment stage (\textit{i.e.,} our first training stage) is to project the visual features into the text input space, and it is the most important step towards making the LLM understand images. In Tab.~\ref{tab:vl_align}, we examine the impact of this stage, and we found that without VL alignment, the performance of all four tasks becomes worse, with the performance of A150-OV and RefCOCO being significantly affected, for example, the A150-OV dropped by -1.1 mAP and RefCOCO even dropped by -4.4 cIoU, probably because these two tasks require a strong requirement on understanding the relationship between vision and language. This result also suggests that the VL alignment is essential, and the LMM-based segmentation models have strong potential.

\noindent\textbf{Joint Training.}
Our architecture design and input schema help integrate various segmentation tasks so that they can be trained on one model. Tab.~\ref{tab:joint} shows the effect of this joint training on different tasks. For the task-specific models, we perform the training on the corresponding task data for 18k iterations. In contrast, the joint training setting (see implementation section for details) has a total of 56k training iterations, which corresponds to 14k iterations per task.

The results show that joint training of different tasks greatly improves the performance of the model. This suggests that learning between tasks is mutually beneficial, which is also the secret of success in LLM. For example, the generic segmentation task helps refine the mask prediction in the referring segmentation. In addition, referring expressions also enhance the model’s ability to recognize more unseen categories, which in turn improves the performance of open-vocabulary segmentation tasks.

\begin{table}[h!]
\centering
\footnotesize
\caption{Comparison with the state-of-the-art methods on three referring image segmentation benchmarks with cIoU. (ft) denotes models further finetuned on RefCOCO/+/g after mix training.
We abbreviate the datasets: COCO(C)~\cite{coco}, LVIS(L)~\cite{lvis}, RefCOCO(RC)~\cite{refcoco}, Object365(O365)~\cite{shao2019objects365}, Video segmentation datasets(V), ADE20K(A)~\cite{ade20k}, COCO-Stuff(CS)~\cite{coco_stuff}, PACO-LVIS(PL)~\cite{paco_lvis}, PASCAL-Part(PP)~\cite{pascal_part}, GranD(G)~\cite{glamm}, VOC2010(VOC)~\cite{pascal-voc-2010}, Visual Genome(VG)~\cite{vg}, Flicker30k(F30K)~\cite{plummer2015flickr30k}, MUSE(M)~\cite{pixellm}, gRefCOCO(gRC)~\cite{grefcoco}, COCO-Interactive(CI).}
\renewcommand{\arraystretch}{0.9}
\scalebox{0.7}{
\begin{tabular}{lcccccccccccc}
\toprule
\multicolumn{1}{c}{\multirow{2}{*}{Method}} & \multirow{2}{*}{Segmentation Data} & \multirow{2}{*}{LLM Type} & \multicolumn{3}{c}{RefCOCO} &  & \multicolumn{3}{c}{RefCOCO+} &  & \multicolumn{2}{c}{RefCOCOg} \\ \cmidrule{4-6} \cmidrule{8-10} \cmidrule{12-13} 
\multicolumn{1}{c}{}                        &                             &                           & val     & testA   & testB   &  & val     & testA    & testB   &  & val           & test         \\ \midrule
SEEM-L~\cite{seem}                                      & C, L, RC                    & -                         & -       & -       & -       &  & -       & -        & -       &  & 65.6          & -            \\
UNINEXT-L~\cite{uninext}                                   & O365, C, RC, V                           & -                         & 80.3    & 82.6    & 77.8    &  & 70.0    & 74.9     & 62.6    &  & 73.4          & 73.7         \\
UNINEXT-H~\cite{uninext}                                   & O365, C, RC, V                            & -                         & 82.2    & 83.4    & 81.3    &  & 72.5    & 76.4     & 66.2    &  & 74.7          & \textbf{76.4}         \\ \midrule
LISA~\cite{lisa}                                        & A, CS, RC, PL, PP                             & Vicuna-7B                 & 74.1    & 76.5    & 71.1    &  & 62.4    & 67.4     & 56.5    &  & 66.4          & 68.5         \\
\gray{LISA(ft)}~\cite{lisa}                                    & \gray{A, CS, RC, PL, PP}                           & \gray{Vicuna-7B}                 & \gray{74.9}    & \gray{79.1}    & \gray{72.3}    &  & \gray{65.1}    & \gray{70.8}     & \gray{58.1}    &  & \gray{67.9}          & \gray{70.6}         \\
GLaMM~\cite{glamm}                                       & G, RC                            & Vicuna-7B                 & 79.5    & 83.2    & 76.9    &  & 72.6    & 78.7     & 64.6    &  & 74.2          & 74.9         \\
u-LLaVA~\cite{ullava}                                     & A, CS, RC, PL, VOC                           & Vicuna-7B                 & 80.4    & 82.7    & 77.8    &  & 72.2    & \textbf{76.6}     & 66.8    &  & \textbf{74.8}          & 75.6         \\
PerceptionGPT~\cite{pi2023perceptiongpt}                               & RC, VG, F30k                           & Vicuna-13B                & 75.3    & 79.1    & 72.1    &  & 68.9    & 74.0     & 61.9    &  & 70.7          & 71.9         \\
PixelLM~\cite{pixellm}                                     & A, CS, RC, PL, M                          & Llama2-13B                & 73.0    & 76.5    & 68.2    &  & 66.3    & 71.7     & 58.3    &  & 69.3          & 70.5         \\
GSVA~\cite{xia2023gsva}                                        & A, CS, RC, PL, PP, gRC                           & Llama2-13B                & 77.7    & 79.9    & 74.2    &  & 68.0    & 71.5     & 61.5    &  & 73.2          & 73.9         \\
\gray{GSVA(ft)}~\cite{xia2023gsva}            & \gray{A, CS, RC, PL, PP, gRC}                           & \gray{Llama2-13B}                & \gray{79.2}    & \gray{81.7}    & \gray{77.1}    &  & \gray{70.3}    & \gray{73.8}     & \gray{63.6}    &  & \gray{75.7}          & \gray{77.0}         \\ \midrule
PSALM                                       & C, RC, CI                           & Phi-1.5 (1.3B)            & \textbf{83.6}    & \textbf{84.7}    & \textbf{81.6}    &  & \textbf{72.9}    & 75.5     & \textbf{70.1}    &  & 73.8          & 74.4         \\ \bottomrule
\end{tabular}
}
\label{tab:ref_coco_system}
\end{table}

\subsection{System-Level Comparison on In-Domain Tasks}
In this section, we compared our model on three in-domain tasks with other state-of-the-arts to illustrate the effectiveness of our approach.

\noindent\textbf{Referring Segmentation.} 
Most existing works aimed at getting LMMs to perform image segmentation are designed for reference segmentation tasks. We compare PSALM with other works on RefCOCO, RefCOCO+, and RefCOCOg, and Tab.~\ref{tab:ref_coco_system} shows the results. 
Owing to the generalized and flexible design of PSALM and the advantages of joint training on multiple tasks and datasets, our system was able to achieve state-of-the-art (SOTA) performance on RefCOCO and RefCOCO+, and competitive performance on RefCOCOg with STOA, despite being driven by LLM with only 1.3B parameters. 
It is worth noting unlike methods such as LISA and GSVA, which may achieve improvements through task-specific fine-tuning (gray-labeled results), PSALM does not perform additional fine-tuning but still achieves better performance on RefCOCO and RefCOCO+ than their fine-tuned models.

\noindent\textbf{Generic Segmentation.} 
We evaluate PSALM with state-of-the-art methods on the COCO Panoptic Segmentation validation set (Tab.~\ref{tab:sys_pan}). Here, we follow the evaluation protocol used in Mask2Former to report PQ, which is the main metric for panoptic segmentation, mAP on thing classes for instance segmentation, and mIoU by merging instance masks from the same category for semantic segmentation. 
Compared to other methods, PSALM achieves comparable performance at similar visual backbone sizes, demonstrating that PSALM is a powerful architecture, even when compared to approaches designed for specific tasks.

\begin{table}[h!]
\centering
\caption{Comparison with the state-of-the-art methods on Panoptic COCO-val. We abbreviate the datasets: COCO-SAM(CM)~\cite{omgseg}, VIPSeg (VIP)~\cite{vipseg}, while others following Tab.~\ref{tab:ref_coco_system}.}
\footnotesize
\renewcommand{\arraystretch}{0.9}
\scalebox{0.85}{
\begin{tabular}{lllclclclc}
\toprule
Method      & Backbone                                &  & Seg. Data                                &  & PQ   &  & mAP   &  & mIoU \\ \midrule
Mask2Former~\cite{mask2former} &  Swin-B       &  &  C             &  & 55.1 &  & 45.2 &  & 65.1 \\
Mask2Former~\cite{mask2former} &  Swin-L       &  &  C             &  & 57.8 &  & 48.6 &  & 67.4 \\
X-Decoder~\cite{xdecode}   &  DaViT-B      &  &  C, L, RC      &  & 56.2 &  & 45.8 &  & 66.0 \\
SEEM~\cite{seem}        &  DaViT-B      &  &  C, L, RC      &  & 56.1 &  & 46.4 &  & 66.3 \\
OMG-Seg~\cite{omgseg}     &  ConvNeXt-XXL &  &  C, VIP, CM, V &  & 55.4 &  & -    &  & -    \\ \midrule
PSALM       &  Swin-B       &  &  C, RC, CI     &  & 55.9 &  & 45.7 &  & 66.6 \\ \bottomrule
\end{tabular}
}
\label{tab:sys_pan}
\end{table}

\begin{table}[h!]
\centering
\caption{Comparison with the state-of-the-art methods on COCO-Interactive. The results of \gray{SEEM-B$^*$} is the result reported in official paper, which is evaluated on 600 random samples of \texttt{COCO-val}, while all others are evaluated on all samples of \texttt{COCO-val}. Abbreviations for each dataset are the same as Tab.~\ref{tab:sys_pan}.}
\footnotesize
\renewcommand{\arraystretch}{0.9}
\scalebox{0.8}{
\begin{tabular}{llclccccccccccc}
\toprule
\multirow{2}{*}{Method}            &  & \multirow{2}{*}{Seg. Data}       &  & \multicolumn{2}{c}{Point}                                &  & \multicolumn{2}{c}{Scribble}                             &  & \multicolumn{2}{c}{Box}                                  &  & \multicolumn{2}{c}{Mask}                              \\ \cmidrule{5-6} \cmidrule{8-9} \cmidrule{11-12} \cmidrule{14-15} 
                                   &  &                                  &  & mIoU                         & cIoU                      &  & mIoU                         & cIoU                      &  & mIoU                         & cIoU                      &  & mIoU                         & cIoU                      \\ \midrule
SAM-B~\cite{sam}                              &  & SA-1B                            &  & 48.7                         & 33.6                      &  & -                            & -                         &  & 73.7                         & 68.7                      &  & -                            & -                         \\
SAM-L~\cite{sam}                               &  & SA-1B                            &  & 51.8                         & 37.7                      &  & -                            & -                         &  & 76.6                         & 71.6                      &  & -                            & -                         \\
SEEM-B~\cite{seem}                             &  & C, L, RC                         &  & 47.8                         & 57.8                      &  & 43.0                         & 44.0                      &  & 44.9                         & 42.1                      &  & 48.4                         & 65.0                      \\
\gray{SEEM-B$^*$}~\cite{seem} &  & \gray{C, L, RC} &  & \gray{81.7} & \gray{-} &  & \gray{83.5} & \gray{-} &  & \gray{75.7} & \gray{-} &  & \gray{76.0} & \gray{-} \\
OMG-Seg~\cite{omgseg}                            &  & C, VIP, CM, V                    &  & 59.3                         & -                         &  & -                            & -                         &  & -                            & -                         &  & -                            & -                         \\ \midrule
PSALM                              &  & C, RC, CI                        &  & 64.3                         & 74.0                      &  & 66.9                         & 80.0                      &  & 67.3                         & 80.9                      &  & 67.6                         & 82.4                      \\ \bottomrule
\end{tabular}
}
\label{tab:inter_coco}
\end{table}

\noindent\textbf{Interactive Segmentation}. 
We also evaluate PSALM in the interactive segmentation tasks. Since the task does not have a well-developed dataset containing all four instructions, previous works have typically used the in-house dataset, thus we re-evaluate other methods on the COCO interactive validation set. The results are shown in Tab.~\ref{tab:inter_coco}, and PSALM achieves leading performance on point, scribble, and mask instructions than all other methods, while on box instruction, SAM performs better on mIoU but worse on cIoU, and we hypothesize might be caused by the different distribution of training data, and the fact that SAM is trained on SA-1B~\cite{sam}, which has a much larger data scale than what we used. 
In addition, we also report the official results of SEEM, which only evaluated 600 samples from the COCO validation set as a reference. 

\subsection{Generalizability on Out-of-Domain Tasks }
Thanks to the flexible design of architecture and input schema, multi-task joint-training, and the strong visual understanding capability of LMM, PSALM shows excellent performance on in-domain tasks, but more importantly, PSALM also demonstrates great potential to generalize to out-of-domain tasks in the zero-shot setting. In this section, we conduct experiments on three different out-of-domain tasks: open-vocabulary segmentation, generalized referring expression segmentation, and video object segmentation. We also tested the zero-shot result of the correspondence benchmark in Ego-Exo4D\cite{egoexo} in the Appendix.

\begin{table}[h!]
\centering
\footnotesize
\caption{Comparison with the state-of-the-art methods on open-vocabulary instance segmentation and semantic segmentation benchmarks. We use mAP for instance segmentation and mIoU for semantic segmentation.We abbreviate the datasets: Pascal Context-459(PC459)~\cite{pc}, Pascal Context-59(PC59)~\cite{pc}, Pascal VOC-20(PAS20)~\cite{pascal-voc-2010}}
\scalebox{0.85} {
\begin{tabular}{lccccccc}
\toprule
\multirow{2}{*}{Method} & \multicolumn{2}{c}{OV Instance Seg.} &  & \multicolumn{4}{c}{OV Semantic Seg.} \\ \cmidrule{2-3} \cmidrule{5-8} 
                        & A150          & Cityscapes         &  & PC459   & A150  & PC59  & PAS20  \\ \midrule
MaskCLIP~\cite{maskclip}               & 6.0             & -                  &  & 10.0     & 23.7   & 45.9   & -       \\
ODISE~\cite{odise}                   & 14.4            & -                  &  & 14.5     & 29.9   & 57.3   & -       \\
SAN~\cite{SAN}                     & 10.6            & -                  &  & 17.1     & 33.3   & 60.2   & 95.5    \\ \midrule
PSALM                   & 9.0             & 20.5               &  & 10.2     & 18.2   & 48.5   & 81.3    \\
PSALM+LVIS              & 13.9            & 19.3               &  & 14.0     & 24.4   & 57.2   & 95.0    \\ \bottomrule
\end{tabular}
}
\label{tab:ov_system}
\end{table}

\noindent\textbf{Open-Vocabulary Segmentation.} 
We first evaluate our PSALM in open-vocabulary segmentation tasks, which require the model to have the ability to deal with unseen categories in training. Here, we conduct experiments on both open-vocabulary instance segmentation and open-vocabulary semantic segmentation, and Tab.~\ref{tab:ov_system} shows the results. Without any special design, PSALM achieves reasonably good performance, although it is still worse than the best specific method in this task, such as SAN, but we believe that PSALM has a strong potential to be further improved by adding more diverse training data, which is advantages of our method. We have also made a preliminary attempt by further involving the LVIS dataset, and as we expected, the performance has significantly improvements.

In addition, existing open-vocabulary segmentation methods are built upon the CLIP model or diffusion model, while our approach is based on LMM models, which is a new path and attempt to bring new inspiration to the community, which we believe is even more important than the performance.

\begin{table}[h!]
\centering
\footnotesize

\caption{Our method's zero-shot performance on gRefCOCO. (ft) denotes models further fine-tuned on gRefCOCO after mix training.}

\renewcommand{\arraystretch}{0.9}
\scalebox{0.8}{
\begin{tabular}{lcccccccccc}
\toprule
\multirow{2}{*}{Methods} & \multirow{2}{*}{LLM Type} & \multirow{2}{*}{Zero-Shot} & \multicolumn{2}{c}{val} &  & \multicolumn{2}{c}{testA} &  & \multicolumn{2}{c}{testB} \\ \cmidrule{4-5} \cmidrule{7-8} \cmidrule{10-11} 
                         &                           &                            & cIoU       & gIoU       &  & cIoU        & gIoU        &  & cIoU        & gIoU        \\ \midrule
MattNet~\cite{yu2018mattnet}                  & -                         & \XSolidBrush                          & 47.5       & 48.2       &  & 58.7        & 59.3        &  & 45.3        & 46.1        \\
LTS~\cite{lts}                      & -                         & \XSolidBrush                          & 52.3       & 52.7       &  & 61.9        & 62.6        &  & 49.9        & 50.4        \\
ReLA~\cite{grefcoco}                    & -                         & \XSolidBrush                          & \textbf{62.4}       & \textbf{63.6}       &  & \textbf{69.3}        & 70.0        &  & 59.9        & 61.0        \\ \midrule
LISA~\cite{lisa}                     & Vicuna-7B                         & \XSolidBrush                          & 38.7       & 32.2       &  & 52.6        & 48.5        &  & 44.8        & 39.7        \\
\gray{LISA(ft)}~\cite{lisa}                     & \gray{Vicuna-7B}                         & \gray{\XSolidBrush}                          & \gray{61.7}       & \gray{63.3}       &  &  \gray{69.2}       & \gray{70.1}        &  & \gray{60.3}        & \gray{61.3}        \\
GSVA~\cite{xia2023gsva}                      & Vicuna-7B                         & \XSolidBrush                          & 61.7       & 63.3       &  & 69.2        & \textbf{70.1}        &  & \textbf{60.3}        & \textbf{61.3}        \\ 
\gray{GSVA(ft)}~\cite{xia2023gsva}                     & \gray{Vicuna-7B}                         & \gray{\XSolidBrush}                          & \gray{63.3}       & \gray{66.5}       &  &  \gray{69.9}       & \gray{71.1}        &  & \gray{60.5}        & \gray{62.2}        \\\midrule
PSALM                     & Phi-1.5 (1.3B)                         & \Checkmark                          & 42.0       & 43.3       &  & 52.4        & 54.5        &  & 50.6        & 52.5        \\
\bottomrule
\end{tabular}
}
\label{tab:grefcoco}
\end{table}

\noindent\textbf{Generalized Referring Expression Segmentation.} 
The referring segmentation datasets used in training contain only a single object, however, the design of the mask proposal allows PSALM to directly address multi-target without any further training or fine-tuning. We evaluate the gRefCOCO benchmark which contains multiple segment targets. In practice, given an expression, we compute the similarity with all mask proposals and retain all masks with similarity greater than 0.6 as foreground. Tab.~\ref{tab:grefcoco} shows the results, PSALM also achieved very promising performance, even outperforming the LISA version that only pre-trained on gRefCOCO but without task-specific fine-tuning.

\begin{table}[h!]
\centering
\footnotesize
\caption{Our method's zero-shot performance on DAVIS-2017 \textit{val}. Note the SEEM report results on 345 randomly sampled frames, while others are evaluated on all frames.}
\renewcommand{\arraystretch}{0.9}
\scalebox{0.85}{
\begin{tabular}{lcccccccc}
\toprule
Methods                        &  & Video Data                                          &  & J\&F                         &  & J                            &  & F                            \\ \midrule
XMem~\cite{cheng2022xmem}                           &  & \Checkmark                           &  & 87.7                         &  & 84.0                         &  & 91.4                         \\
OMG-Seg~\cite{omgseg}                        &  & \Checkmark                           &  & 76.9                         &  & -                            &  & -                            \\
Painter~\cite{painter}                        &  & \XSolidBrush                         &  & 34.6                         &  & 28.5                         &  & 40.8                         \\
SegGPT~\cite{wang2023seggpt}                         &  & \XSolidBrush                         &  & 75.6                         &  & 72.5                         &  & 78.6                         \\
\gray{SEEM-B}~\cite{seem} &  & \gray{\XSolidBrush} &  & \gray{62.8} &  & \gray{59.5} &  & \gray{66.2} \\ \midrule
PSALM                          &  & \XSolidBrush                         &  & 68.8                         &  & 65.9                         &  & 71.7                         \\ \bottomrule
\end{tabular}
}
\label{tab:davis}
\end{table}

\noindent\textbf{Video Object Segmentation.} 
We also evaluate PSALM on video object segmentation task, which typically gives an image of the first frame and a segmentation reference, and expect to segment subsequent videos based on the given reference. DAVIS-2017~\cite{davis} is the most commonly used dataset, and we test PSALM on it. During inference, we extract the region feature based on the last-frame prediction (or first-frame segmentation reference) as the visual-prior condition and use the image of the current frame to predict the mask. Tab.~\ref{tab:davis} shows the results, without training on any video data, PSALM shows promising zero-shot performance. 

\section{Conclusion}
The PSALM proposed in this study extends the capability of LMM from text output tasks to image segmentation, addresses the output limitations of the LMM, and unifies various segmentation tasks with a novel input model. PSALM exhibits excellent performance in multiple in-domain tasks, and its generalization ability in out-of-domain tasks further underscores its potential. The success of PSALM in in-domain and cross-domain tasks highlights the importance of flexibility and adaptability in model design, paving the way for future innovations in visual understanding.

\bibliographystyle{splncs04}
\bibliography{main}
\newpage
\section{Prompts for Different Tasks}
As discussed in Sec.~\ref{sec:condition_prompt}, our input schema has three kinds of inputs: task instruction prompt, condition prompt, and a set of mask tokens. 
For different tasks, we use different instruction prompts and condition prompts, as listed in Tab.~\ref{tab:prompts}. 
Basically, tasks that use the same type of condition prompts also adopt the same task instruction prompt.

In addition, some tasks that use category condition often need to deal with background as well, such as instance segmentation, for which we specifically append a \textit{`background`} at the end of the joint sentence. 
\setcounter{table}{13}
\begin{table}[h!]
\footnotesize
\centering
\caption{Detail prompts for all tasks.}
\scalebox{0.7}{
\begin{tabular}{c|c|c|c}
\toprule
Task                       & Dataset                                                 & Instruction prompt                                                                      & Condition prompt                                                                                 \\ \midrule
\multirow{2}{*}{Panoptic Seg.}              & \multirow{2}{*}{COCO}                                   & \multirow{2}{*}{\textit{\makecell[c]{You need to segment all objects.\\ This is all the candidate categories:}}} & \multirow{2}{*}{\texttt{[class1], [class2] ...}}                                               \\
                           &                                                         &                                                                                         &                                                                                                  \\
& & & \\
                           
\multirow{2}{*}{OV Seg.}                    & \multirow{2}{*}{ADE20K, \textit{etc.}} & \multirow{2}{*}{\textit{\makecell[c]{You need to segment all objects.\\ This is all the candidate categories:}}} & \multirow{2}{*}{\texttt{[class1], [class2], ...}}                                                \\
                           &                                                         &                                                                                         &                                                                                                  \\
& & & \\
\multirow{2}{*}{Referring Seg.}             & \multirow{2}{*}{RefCOCO/+/g}                            & \multirow{2}{*}{\textit{\makecell[c]{Please segment according to \\ the following instruction:}}}                 & \multirow{2}{*}{\texttt{object description}}                                              \\
                           &                                                         &                                                                                         &                                                                                                  \\
& & & \\
\multirow{2}{*}{Generalized Referring Seg.} & \multirow{2}{*}{gRefCOCO}                               & \multirow{2}{*}{\textit{\makecell[c]{Please segment according to \\ the following instruction:}}}                  & \multirow{2}{*}{\texttt{object description}}                                      \\
                           &                                                         &                                                                                         &                                                                                                  \\
& & & \\
\multirow{2}{*}{Interactive Seg.}           & \multirow{2}{*}{COCO-Interactive}                       & \multirow{2}{*}{\textit{\makecell[c]{Please segment by given regions:}}}                                       & \multirow{2}{*}{\texttt{<interaction1>, <interaction2>...}} \\
                           &                                                         &                                                                                         &                                                                                                  \\
& & & \\
\multirow{2}{*}{Video Object Seg.}          & \multirow{2}{*}{DAVIS}                                  & \multirow{2}{*}{\textit{\makecell[c]{Please segment by given regions:}}}                                    & \multirow{2}{*}{\texttt{<interaction1>, <interaction2>...}} \\
                           &                                                         &                                                                                         &                                                                                                  \\
                    & & & \\
        \multirow{2}{*}{Ego-exo Correspondence}                   & \multirow{2}{*}{Ego-Exo4D}                              & \multirow{2}{*}{\textit{\makecell[c]{Please segment by given regions:}}}                                    & \multirow{2}{*}{\texttt{<interaction1>, <interaction2>...}} \\
     &                                                         &                                                                                         &                                                                                                  \\ \bottomrule
\end{tabular}
}
\label{tab:prompts}
\end{table}

\section{Training Details}
In this section, we introduce the training details of our model.

Our model has two training stages, the first stage is the vision-language alignment stage, in which we strictly follow the default settings of LLaVA, with the only change being the adoption of Phi-1.5 as the LLM and the use of Swin as the visual encoder with the hyper-parameters as shown in Tab.~\ref{tab:first_train}.

In the second stage, to train the final model, we use a total of 56k training iterations. In each iteration, we randomly sample one task with equal probability from four tasks: generic segmentation (COCO Panoptic), referring segmentation (RefCOCO/+/g), interactive segmentation (COCO-Interactive), and a visual-language instruction task (LLaVA1.5 training data) and all images are resized to $1024^2$ by padding the shorter side to keep the aspect ratio. During training, the visual encoder is frozen while all other model parts are trainable. In addition, we adopt the Hungarian matching to automatically assign the ground-truth of mask proposals during training. In generic segmentation tasks, we use both classification loss and mask loss as cost matrix for matching, while other tasks use only mask loss. Tab.~\ref{tab:second_train} shows the hyper-parameters. 

For ablation, we use the same training setting as the final model, but with the shorter 9k training iterations to reduce the cost of the experiment.

\begin{table}[h!]
\setlength{\tabcolsep}{2.5mm}
\centering
\footnotesize
\caption{Hyper parameters of our model in the first stage training.}
\begin{tabular}{lcccr}
\toprule
Parameters   &&& & Value \\ \midrule
Optimizer    &&& &  AdamW     \\
Learning Rate &&& &  $2\times10^{-3}$     \\
Batch Size &&& &  128    \\
Number of Iteration &&& &  4,650    \\
Learning Rate Schedule &&& &  Cosine Decay     \\
Weigth Decay &&& &  0.0     \\
Warmup Steps &&& &  140     \\
$\beta_1$ &&& &  0.9     \\
$\beta_2$ &&& &  0.999     \\
        Training Data &&& &  CC3M     \\

Image Size &&& &  $1024\times 1024$     \\
\multirow{2}{*}{Image Processing} &&& &  \multirow{2}{*}{\makecell[r]{Resize long edge to 1024 \\ and padding short edge to 1024.}}     \\
        &&&& \\
        \bottomrule
\end{tabular}
\label{tab:first_train}
\end{table}
\begin{table}[h!]
\setlength{\tabcolsep}{0.6mm}
\centering
\footnotesize
\caption{Hyper parameters of our model in the second stage training.}
\begin{tabular}{lcccr}
\toprule
Parameters   &&& & Value \\ \midrule
Optimizer    &&& &  AdamW     \\
Learning Rate &&& &  $4\times10^{-5}$     \\
Batch Size &&& &  64    \\
Number of Iteration &&& &  56,000    \\
Number of Iteration (for Ablation) &&& &  9,000    \\
Learning Rate Schedule &&& &  Cosine Decay     \\
Weigth Decay &&& &  0.0     \\
Warmup Steps &&& &  1680     \\
Warmup Steps (for Ablation) &&& &  270    \\
$\beta_1$ &&& &  0.9     \\
$\beta_2$ &&& &  0.999     \\
        \multirow{2}{*}{Training Data} &&& &  \multirow{2}{*}{\makecell[r]{COCO-Panoptic (25\%); RefCOCO/+/g(25\%); \\ COCO-Interactive(25\%); LLaVA 1.5(25\%)}}     \\

        &&&& \\
Image Size &&& &  $1024\times 1024$     \\
\multirow{2}{*}{Image Processing} &&& &  \multirow{2}{*}{\makecell[r]{Resize long edge to 1024 \\ and padding short edge to 1024.}}     \\
        &&&& \\
        \bottomrule
\end{tabular}
\label{tab:second_train}
\end{table}

\section{Details on building COCO-Interactivate Dataset}
In this section, we describe in detail how to build the COCO-Interactive dataset.
The COCO-Interactivate is based on image and annotations of COCO2017 instance segmentation, which provide the masks and bounding boxes for each instance, and we use the annotations to automatically generate four types of visual prompts: point, scribble, mask, and box. Fig.~\ref{fig:interactive} shows the illustration of the visual prompts, and we will introduce each of them in the following:

\noindent\textbf{Point}. For each instance, we generate a point visual prompt by randomly sampling a point within a circular region centered on the bounding box with a radius of half the short side of the bounding box.  

\noindent\textbf{Scribble}. The generation process of scribble has two steps. First, we randomly jitter the width and height of the ground-truth bounding box from a scale factor range of [0.5, 1.2], and ensure that the IoU between the jittered box with the original box is greater than 0.5. Then, given a jittered box, we randomly select one of its diagonals and generate a sin curve along it. The amplitude of the sin curve is randomly chosen from [10, 20], the frequency is randomly sampled from \([0.2 \times 2\pi, 2\pi]\), and the phase shift is randomly sampled from \([0, 2\pi]\). 

\noindent\textbf{Box}. We randomly jittered ground truth boxes as box prompts, and the length and width of each jittered box were obtained by scaling from a scale sampled in the range [0.9, 1.1].

\noindent\textbf{Mask}. The mask visual prompts are obtained by applying a Gaussian filter on the ground truth mask at first, with the standard deviation of the Gaussian kernel set to 2, and then binarizing the blurred mask. 

\setcounter{figure}{3}

\begin{figure}[!t]
  \centering
  \includegraphics[width=1.\textwidth]{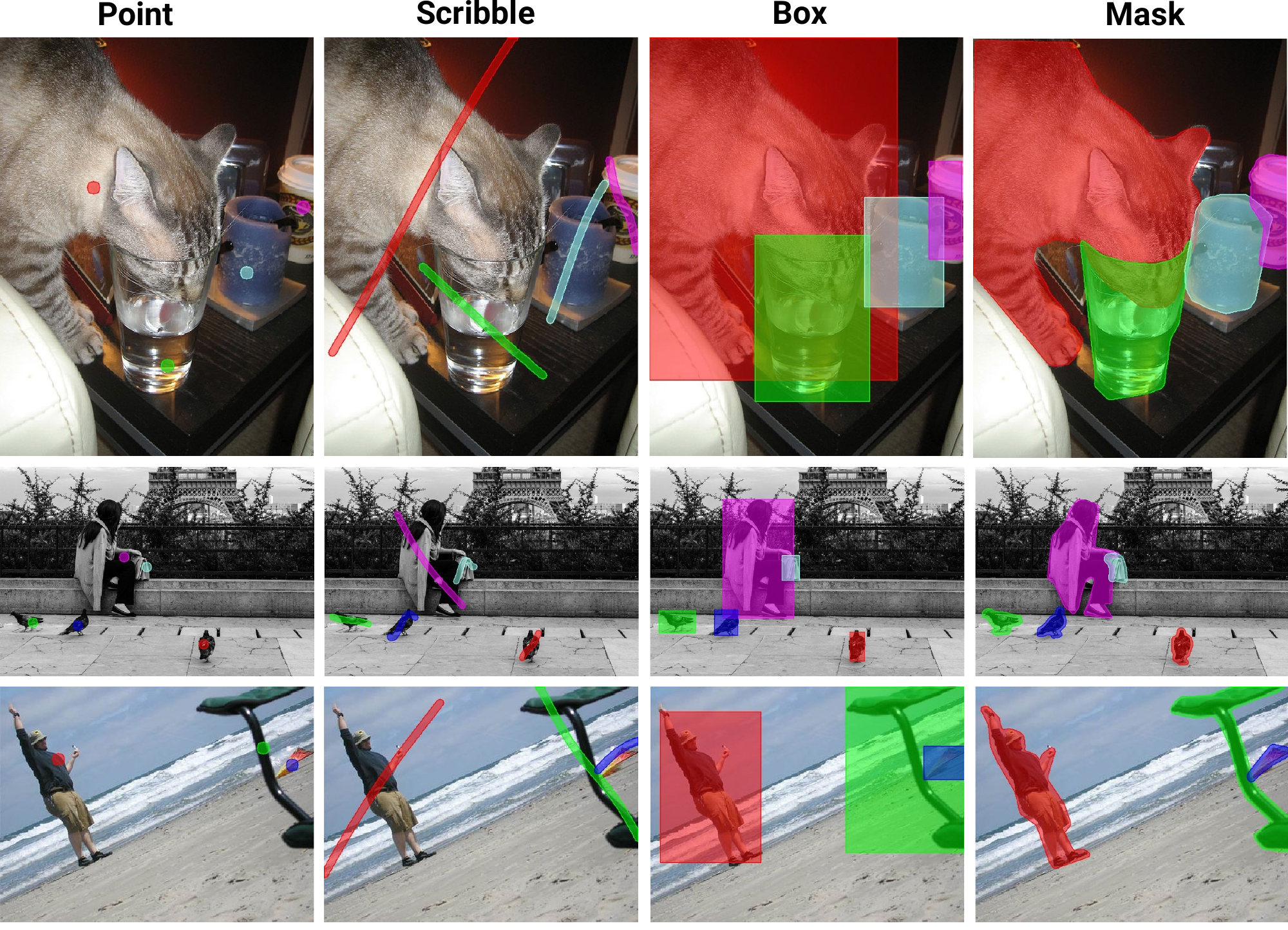}
  \caption{Visualization of different types of visual prompts}
  \label{fig:interactive}
\end{figure}
\section{Datasets}
\noindent\textbf{COCO-Panoptic.} The COCO-Panoptic dataset is an extension of the COCO dataset, specifically designed for panoptic segmentation tasks. It consists of over 200,000 images with detailed annotations that cover 80 object categories for instance segmentation and additional categories for semantic segmentation. 

\noindent\textbf{RefCOCO.} RefCOCO is a dataset designed for the task of referring expression comprehension and segmentation. It consists of images from the COCO dataset that are annotated with referring expressions, where each expression uniquely identifies a particular object within the image. It includes three splits: RefCOCO, RefCOCO+, and RefCOCOg, each with different characteristics and annotation styles.

\noindent\textbf{LVIS.} The LVIS dataset is a benchmark for instance segmentation with a large vocabulary of object categories. It features high-quality instance annotations for over 1,000 object categories across a diverse set of images. LVIS is particularly known for its long-tail distribution of categories, which presents a unique challenge for segmentation models.

\noindent\textbf{ADE20K.} ADE20K is a widely used dataset as an open-vocabulary segmentation benchmark, it contains both things and stuffs annotations and thus can evaluate panoptic segmentation. It encompasses a diverse collection of images from various indoor and outdoor scenes. It is part of the MIT Scene Parsing Benchmark and provides dense pixel-wise annotations for 150 object categories, facilitating research in scene understanding and segmentation.

\noindent\textbf{Cityscapes.} 
Cityscape is a dataset focused on urban street scenes. The dataset contains a large number of high-quality video sequences and pixel-accurate annotations from 30 categories in 50 different urban street scenes. With its detailed instance-level annotations, Cityscapes is pivotal to advancing semantic and instance segmentation research, especially in autonomous driving and urban scene perception.

\noindent\textbf{Pascal VOC.} Pascal VOC contains 20 classes of semantic segmentation annotation.

\noindent\textbf{Pascal Context.} Pascal Context is an extension of the PASCAL VOC dataset, providing comprehensive scene understanding through detailed semantic labels for the entire scene in each image. It comes in two versions: PC-59, which focuses on the most frequent 59 categories, and PC-459 includes a broader set of 459 categories.

\noindent\textbf{DAVIS-2017.} DAVIS-2017 is a video segmentation benchmark that provides high-quality, full-resolution video sequences with per-pixel annotations of multiple objects. It is commonly used to evaluate the performance of video object segmentation methods, particularly in semi-supervised settings where the first-frame mask is provided.

\noindent\textbf{gRefCOCO.} gRefCOCO is the first large-scale Generalized Referring Expression Segmentation dataset that contains multi-target, no-target, and single-target expressions. 

\noindent\textbf{Ego-Exo4D.} Ego-Exo4D is a diverse, large-scale multimodal multiview video dataset and benchmark challenge. Ego-Exo4D centers around simultaneously captured and time-synced egocentric and exocentric vides of skilled human activities. More than 800 participants from 13 cities worldwide performed these activities in 131 different natural scene contexts, yielding long-form captures from 1 to 42 minutes each and 1,422 hours of video combined. The Correspondence benchmark needs the model to predict the corresponding mask for the same object in each synchronized frame of the other view if it is visible.

\noindent\textbf{CC3M.} CC3M is a large-scale dataset of image-caption pairs designed for training and evaluating visual-language models. It contains around three million images sourced from the web, each accompanied by a descriptive caption. 

\section{Implement Details of Decouple Ablation}
In Tab.~3, we compare the decouple design and non-decouple design on COCO Semantic Segmentation. Here, we will introduce the implementation detail for the non-decouple design. Specifically, we omitted the mask token, and instead of using the category condition embeddings as the mask query, and fed into the mask generator to generate masks. In this case, the matching loss mechanism is unnecessary, the condition embedding of a category is used to predict the class and mask at the same time. 

\begin{table}[h!]
\centering
\footnotesize
\caption{Performance on multi-modal benchmarks.}
\begin{tabular}{lccccc}
\toprule
Methods         & LLM Type          & VQA$^{V2}$ & SQA & MMB & POPE \\ \midrule
OpenFlamingo~\cite{flamingo}    & MPT-7B        & 51.8   & -         & 5.7     & -    \\
Kosmos-2~\cite{kosmos_2}        & -            & 51.1   & -         & -       & -    \\
BLIP-2~\cite{blip2}          & Vicuna-13B    & 41.0   & 61.0      & -       & 85.3 \\
InstructionBLIP~\cite{dai2024instructblip} & Vicuna-7B    & -      & 60.5      & -       & 36.0 \\
IDEFICS~\cite{idefics}           & Llama-7B    & 50.9      & 44.2         & 48.2    & - \\
LLaVA-1.5~\cite{llavav1_5}       & Vicuna-7B & 78.5   & 66.8      & 64.3    & 85.9\\
\midrule
PSALM            & Phi-1.5 (1.3B) & 62.3   & 64.9      & 52.5    & 80.3 \\ \bottomrule
\end{tabular}
\label{tab:vl_bench}
\end{table}

\begin{table}[h!]
\caption{Zero-shot performance of Correspondence benchmark on Ego-Exo4D.}
  \begin{minipage}{.48\textwidth}
  \centering
\footnotesize
\scalebox{0.9}{
  \begin{tabular}{llcccc}
\toprule
\multirow{2}{*}{Query Mask} & \multirow{2}{*}{Method} & \multirow{2}{*}{Zero-Shot}  & Test &  & Val \\
                            &                         &                             & IoU  &  & IoU \\ \midrule
Ego                         & XSegTx                 & \Checkmark   & 0.6  &  & -   \\
Ego                         & XMem                    & \Checkmark   & 4.6  &  & -   \\
Ego                         & XSegTx                  & \XSolidBrush & 13.9 &  & -   \\
Ego                         & XMem                    & \XSolidBrush & 14.6 &  & -   \\
\midrule
Ego                         & PSALM                   & \Checkmark   & -    &  & 7.9 \\  \bottomrule
\end{tabular}
}
\label{tab:mask_query}

  \end{minipage}\hfill
  \begin{minipage}{.48\textwidth}
  \centering
  \footnotesize
  \scalebox{0.9}{
  \begin{tabular}{llcccc}
\toprule
\multirow{2}{*}{Query Mask} & \multirow{2}{*}{Method} & \multirow{2}{*}{Zero-Shot}  & Test &  & Val \\
                            &                         &                             & IoU  &  & IoU \\ \midrule
Exo                         & XSegTx                  & \Checkmark   & 1.6  &  & -   \\
Exo                         & XMem                    & \Checkmark   & 21.8 &  & -   \\
Exo                         & XSegTx                  & \XSolidBrush & 43.8 &  & -   \\
Exo                         & XMem                    & \XSolidBrush & 43.4 &  & -   \\
\midrule
Exo                         & PSALM                   & \Checkmark   & -    &  & 9.6 \\ \bottomrule
\end{tabular}
}

  \end{minipage}
\label{tab:ego4d}
\end{table}
\section{Additional Experiment}
\subsubsection{Multi-modal Benchmark Evaluation.} 

Our PSALM model is based on MLLM and thus able to deal with vision and language tasks, and therefore we evaluate our model on several commonly used multi-modal benchmarks, and results are shown in Tab.~\ref{tab:vl_bench}. PSALM achieved promising results compared with other MLLM methods, such as BLIP-2~\cite{blip2} and InstructionBLIP~\cite{dai2024instructblip}. Although our model still lags behind the official LLaVA1.5 7B model, we believe that increasing the model size can largely close the performance gap.

Given the absence of training data, our related works such as LISA\cite{lisa} and PixelLM\cite{pixellm}, despite their theoretical capability to handle such tasks, yield suboptimal results. Take LISA as an instance, it achieves a mere \textbf{0.12} on the VQA score in a zero-shot manner.

\subsubsection{Ego-Exo4D Correspondence Benchmark.} 
\begin{figure}[!t]
  \centering
  \includegraphics[width=1.\textwidth]{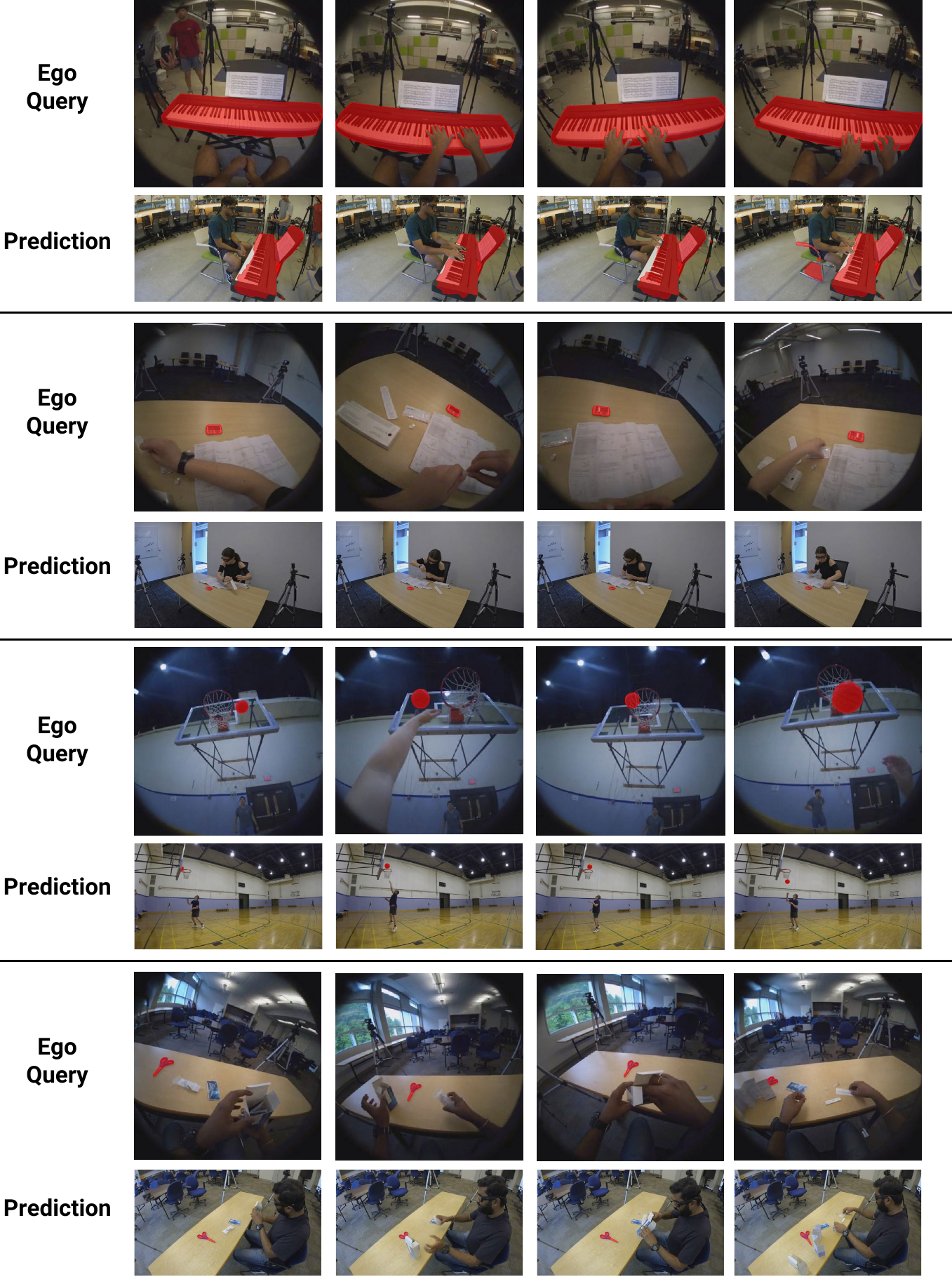}
  \caption{More examples of Ego-exo correspondence in Ego-Exo4D~\cite{egoexo}.}
  \label{fig:Ego-Exo4D}
\end{figure}

Ego-Exo4D~\cite{egoexo} is a large-scale multi-modal multiview video dataset, and its correspondence benchmark is designed to predict the mask of an object in a novel view based on a given view. We evaluate this benchmark in a zero-shot manner to show the task generality of our model for such tasks. 
We performed the evaluation on Ego-Exo4D benchmark, since the official test set and model have not been released yet, we cannot directly compare the performance under the same setting, so we only report the quantitative results on the validation set as a reference in Tab.~\ref{tab:ego4d} and shows more qualitative results in Fig.~\ref{fig:Ego-Exo4D}. 

\section{More Qualitative Results}
\begin{figure}[!t]
  \centering
  \includegraphics[width=1.\textwidth]{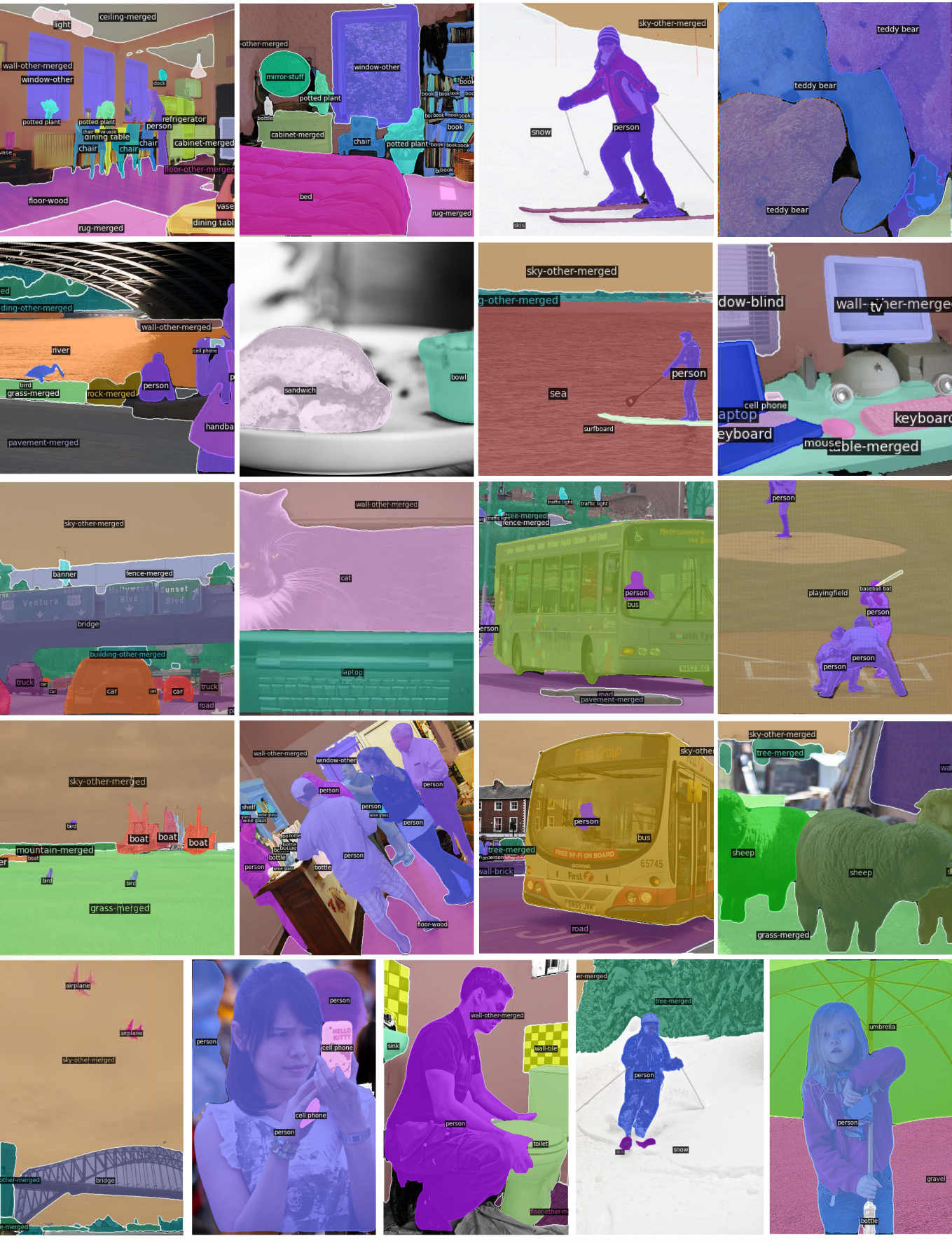}
  \caption{More examples of panoptic segmentation in COCO~\cite{coco}.}
  \label{fig:panoptic}
\end{figure}

\begin{figure}[!t]
  \centering
  \includegraphics[width=1.\textwidth]{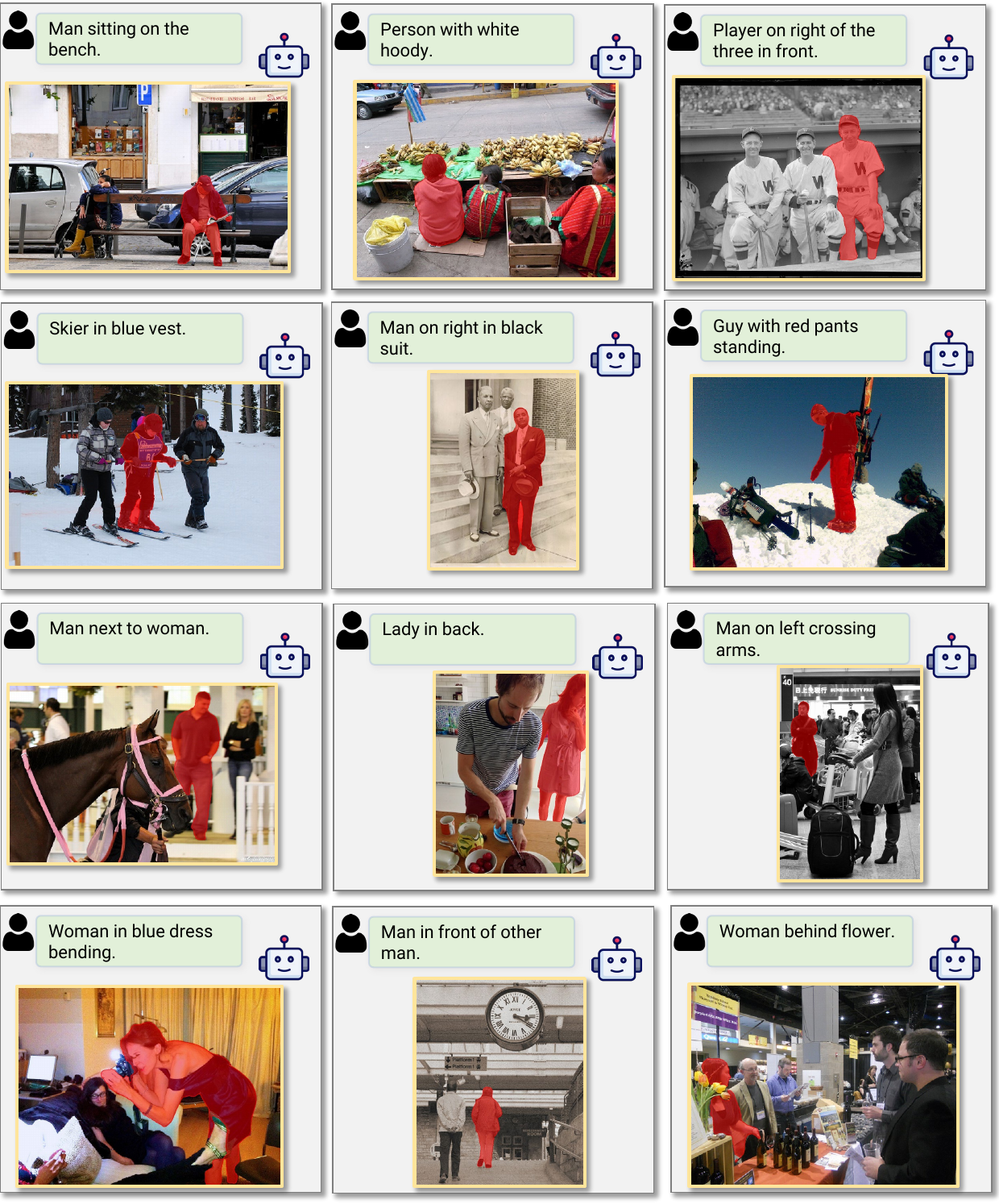}
  \caption{More examples of referring segmentation in RefCOCO~\cite{refcoco}.}
  \label{fig:refer}
\end{figure}

\begin{figure}[!t]
  \centering
  \includegraphics[width=1.\textwidth]{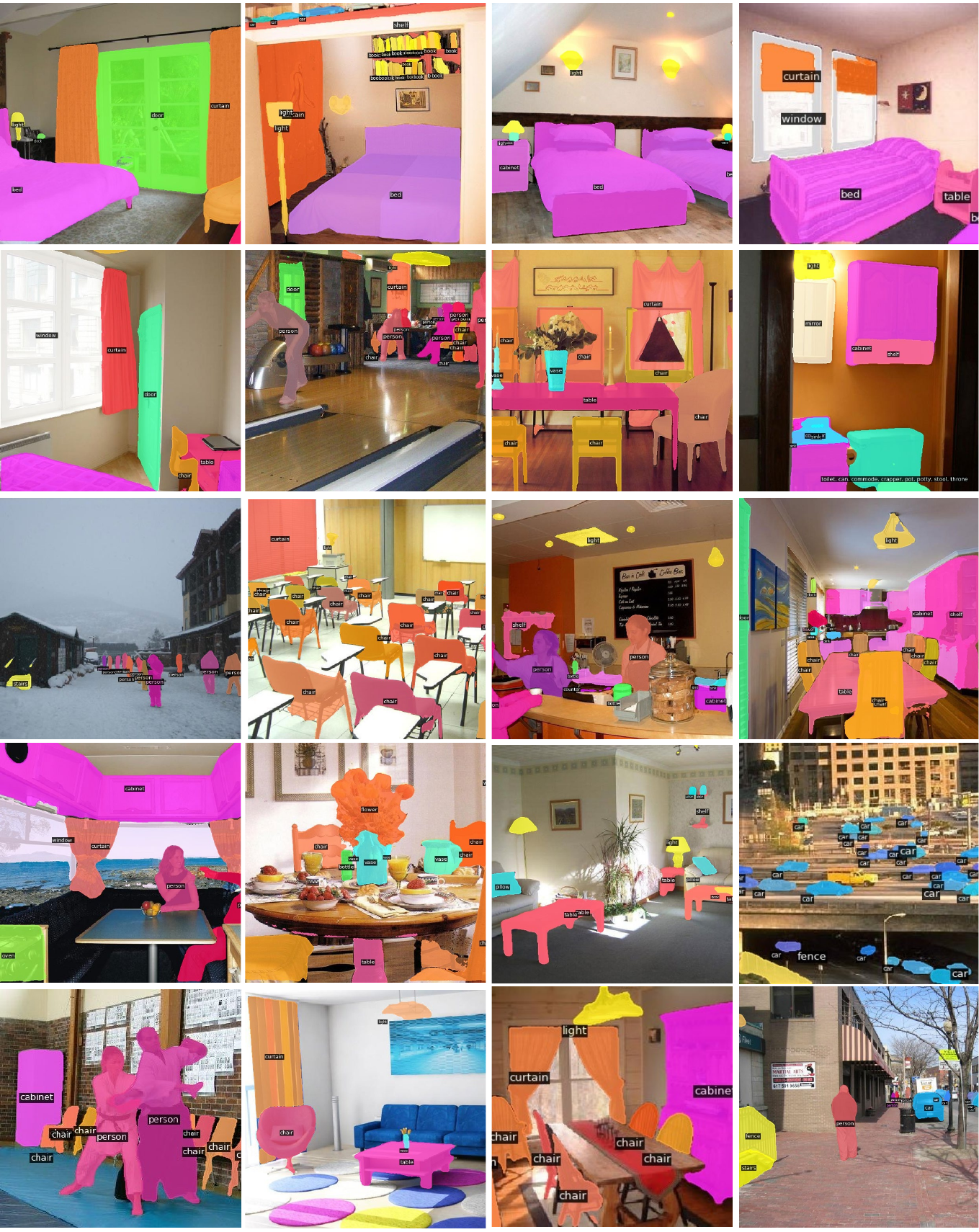}
  \caption{More examples of open-vocabulary instance segmentation on ADE20K~\cite{ade20k}.}
  \label{fig:ov}
\end{figure}

\begin{figure}[!t]
  \centering
  \includegraphics[width=1.\textwidth]{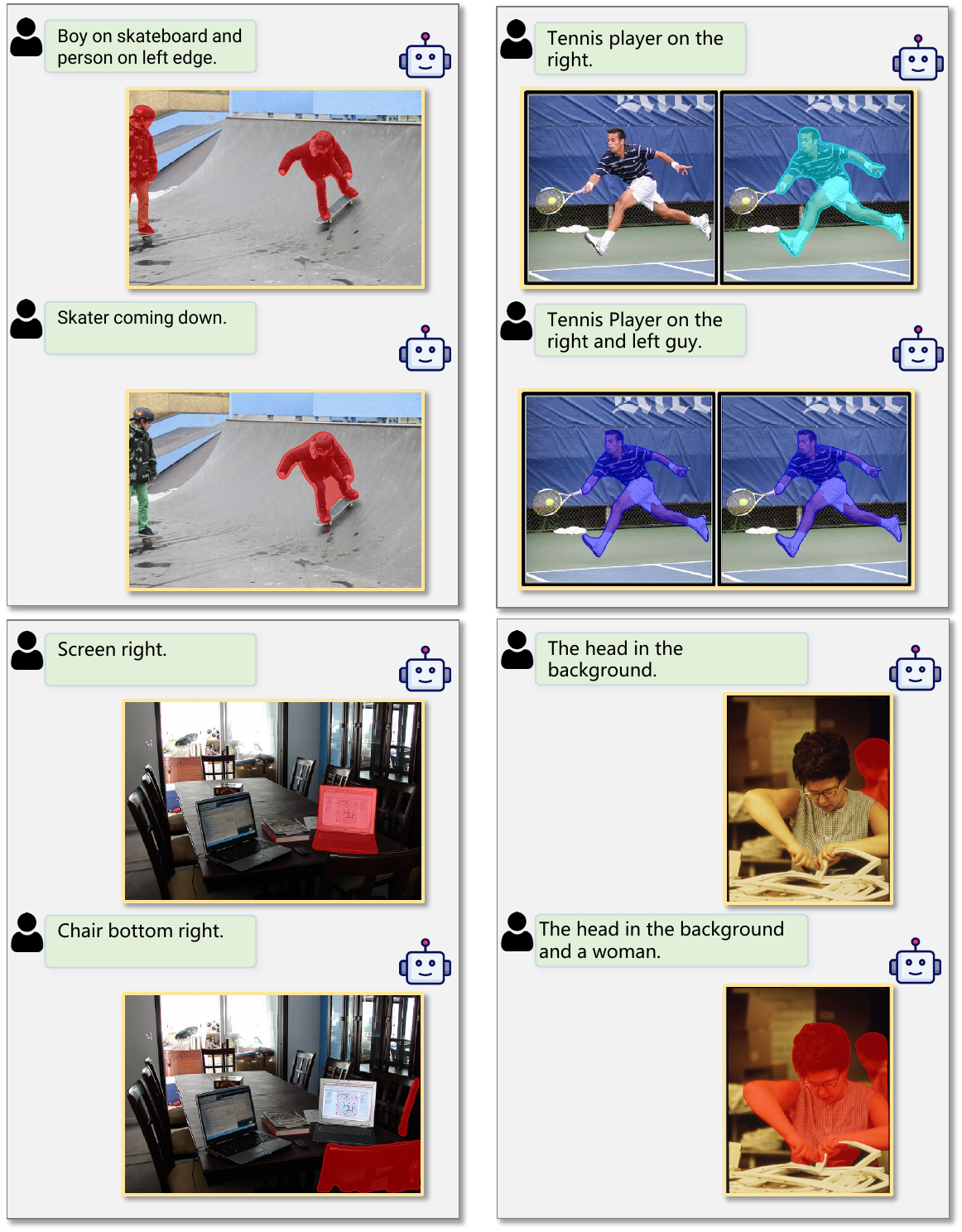}
  \caption{More examples of generalized referring segmentation in gRefCOCO~\cite{grefcoco}.}
  \label{fig:grefcoco}
\end{figure}

Fig.~\ref{fig:panoptic}, Fig.~\ref{fig:refer} and Fig.~\ref{fig:interactive} show more qualitative examples of in-domain tasks. Fig.~\ref{fig:ov}, Fig.~\ref{fig:grefcoco} and Fig.~\ref{fig:davis} shows more examples of out-of-domain tasks.

\begin{figure}[!t]
  \centering
  \includegraphics[width=1.\textwidth]{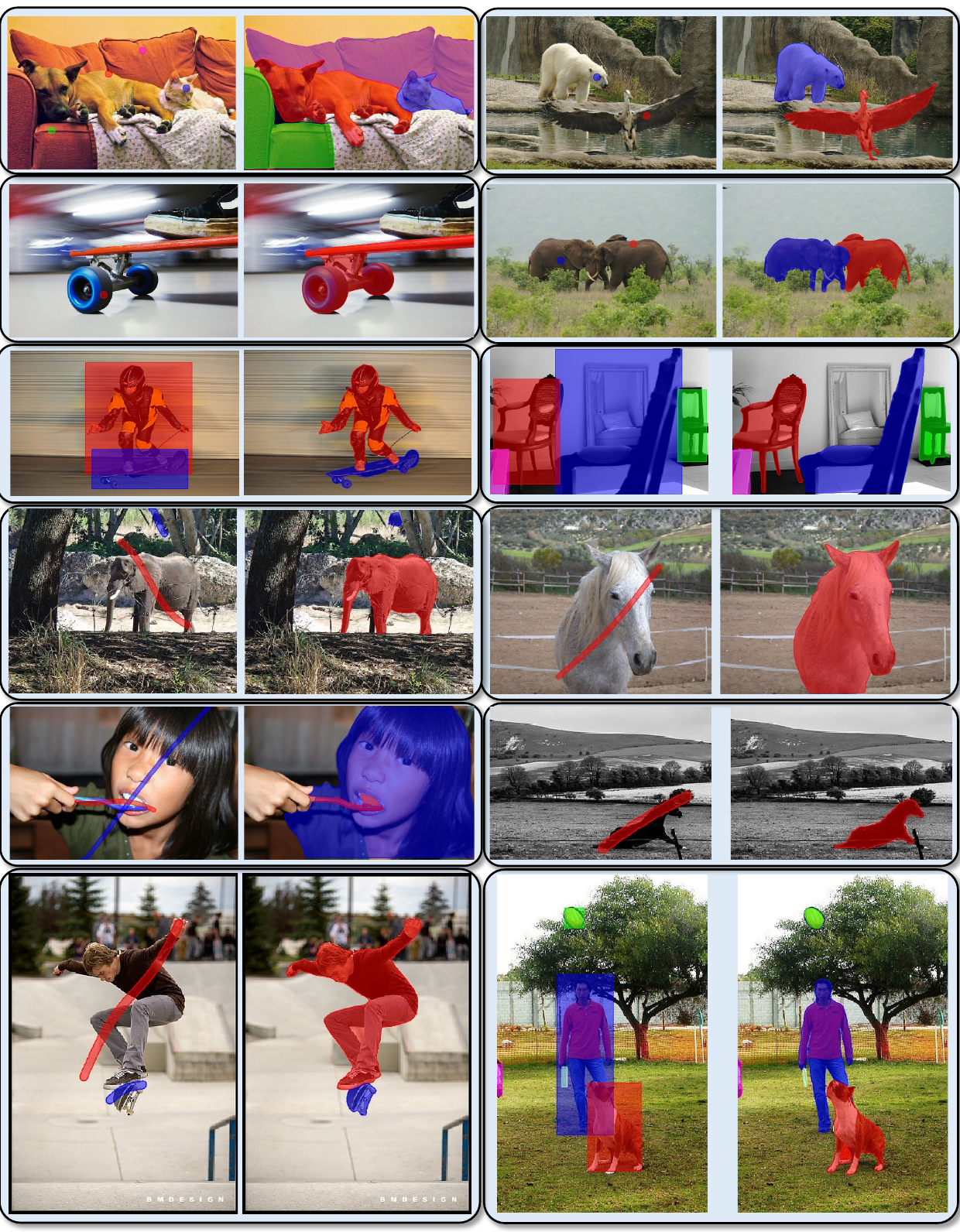}
  \caption{More examples of interactive segmentation in COCO-Interactive.}
  \label{fig:interactive}
\end{figure}
\begin{figure}[!t]
  \centering
  \includegraphics[width=1.\textwidth]{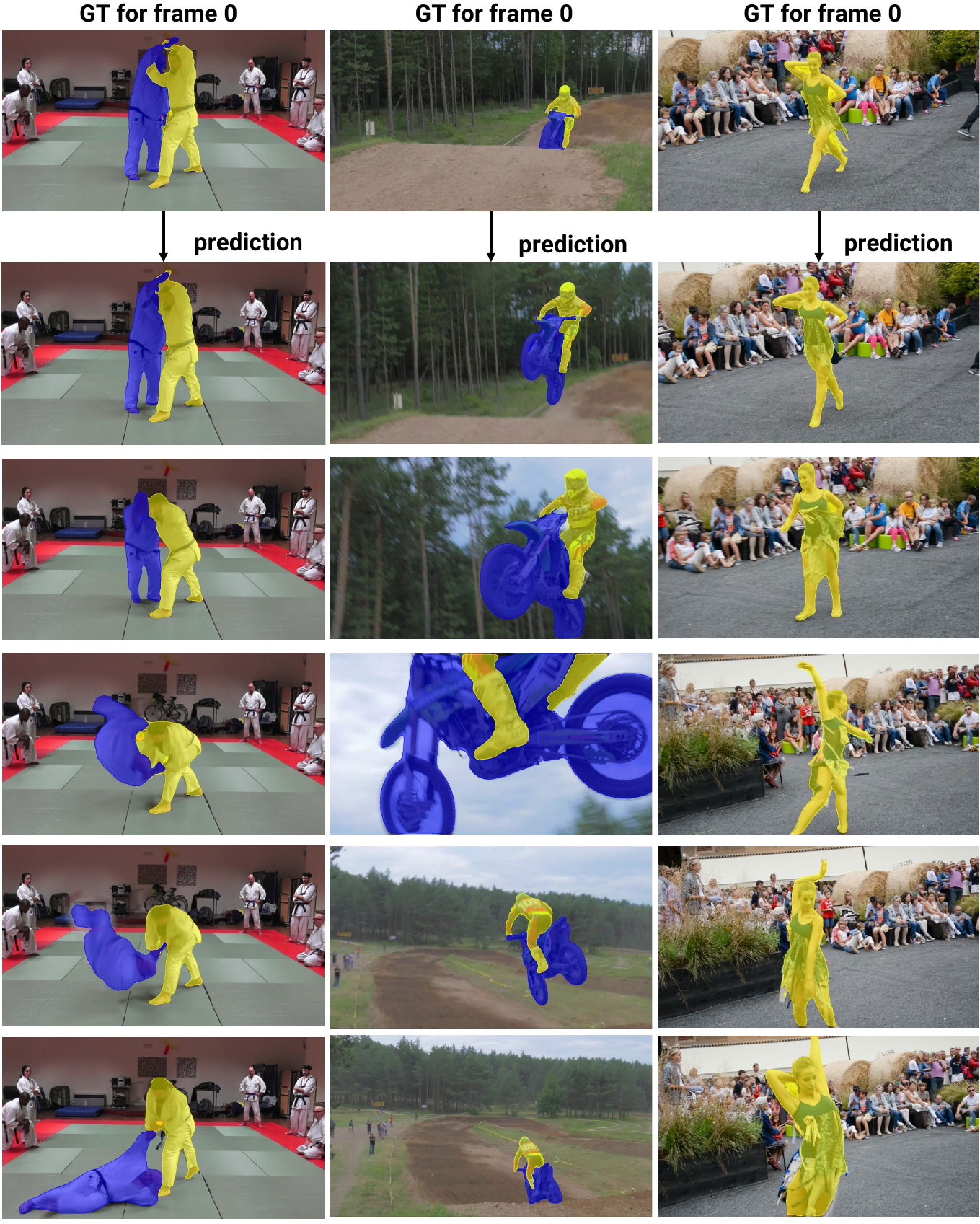}
  \label{fig:davis}
\end{figure}
\begin{figure}[!t]
  \centering
  \includegraphics[width=1.\textwidth]{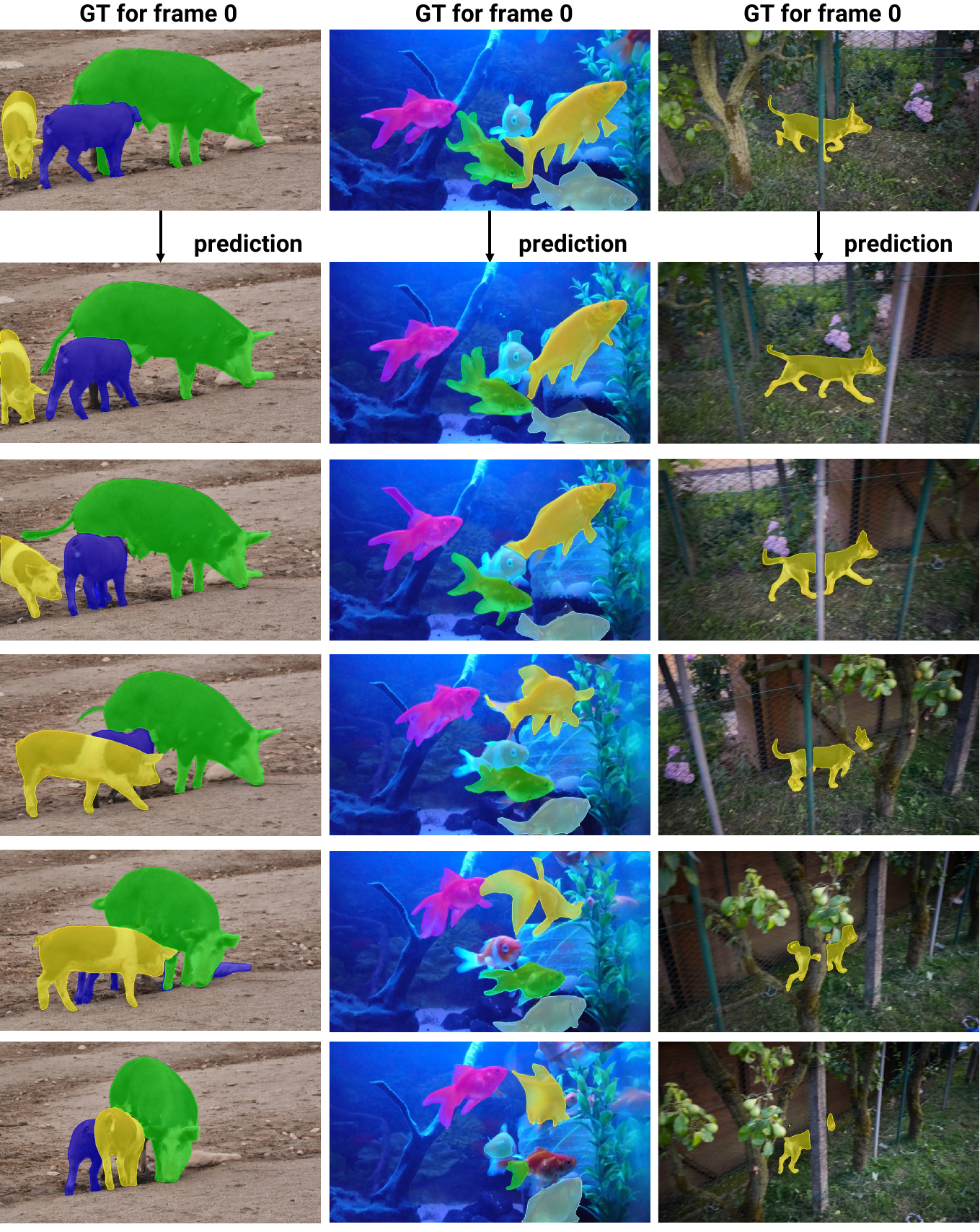}
  \caption{More examples of video object segmentation in DAVIS \textit{val}~\cite{davis}.}
  \label{fig:davis}
\end{figure}
\end{document}